\documentclass[11pt]{article}

\usepackage[preprint]{acl}

\usepackage{times}
\usepackage{latexsym}

\usepackage[T1]{fontenc}

\usepackage[utf8]{inputenc}

\usepackage{microtype}

\usepackage{inconsolata}

\usepackage{graphicx}
\usepackage{algorithm}
\usepackage{algorithmic}
\usepackage{booktabs}
\usepackage{multicol}
\usepackage{amsmath}
\usepackage{amssymb}
\usepackage{multirow} 
\usepackage{tikz} 
\usepackage{pgfplots}
\usepackage{color}
\usepackage{pgfplotstable}
\usepackage{longtable}
\usepackage{tabularx}
\usepackage[dvipsnames,table]{xcolor}
\usepackage{soul}

\usepackage[framemethod=TikZ]{mdframed}
\mdfdefinestyle{hypobox}{
    backgroundcolor=gray!15, 
    linecolor=gray!15,       
    innerleftmargin=10pt,
    innerrightmargin=10pt,
    innertopmargin=10pt,
    innerbottommargin=10pt,
    roundcorner=5pt          
}

\pgfplotsset{compat=1.18}

\usetikzlibrary{patterns}

\definecolor{gray1}{rgb}{0.93, 0.93, 1.0}
\definecolor{darkgreen}{RGB}{1,50,32}
\definecolor{ForestGreen}{RGB}{34,139,34}
\definecolor{blue1}{HTML}{f1eef6} 
\definecolor{blue2}{HTML}{bdc9e1} 
\definecolor{blue3}{HTML}{74a9cf} 
\definecolor{blue4}{HTML}{2b8cbe}

\title{Revealing Algorithmic Deductive Circuits for Logical Reasoning } 

\author{
 \textbf{Phuong Minh Nguyen}\and
 \textbf{Tien Huu Dang}\and
 \textbf{Naoya Inoue} 
\\
  Japan Advanced Institute of Science and Technology 
\\
\texttt{\{phuongnm,tiendh,naoya-i\}@jaist.ac.jp}
\\
}

\begin{document}
\maketitle




\begin{abstract} 

Recent studies have shown that Large Language Models (LLMs) can achieve strong reasoning performance by incorporating functional symbolic representations that abstractly describe \textit{graph traversal algorithms} and \textit{step-by-step reasoning} in few-shot learning settings.   
However, it remains unclear how LLMs genuinely understand the abstract meaning of each reasoning step and the overall algorithm from only a limited number of demonstrations.
This work aims to \textit{localize the attention heads responsible for individual reasoning steps and characterize the types of information transferred among them.} We first align constituent reasoning steps with their corresponding token logits under a symbolic-aided Chain-of-Thought (CoT) prompting framework. Our analysis shows that token positions that steer the reasoning process are associated with low confidence scores caused by constraints on satisfying reasoning behavior patterns in demonstrations. We then adopt causal mediation analysis techniques to identify the attention heads responsible for these patterns.  In addition, our findings indicate that LLMs retrieve factual and rule-based information for individual sub-reasoning tasks through specialized attention heads (approximately 3\% total heads), whereas higher layers predominantly facilitate information integration and the emergence of global reasoning strategies (e.g., graph traversal algorithms) that coordinate multiple intermediate reasoning steps to solve the overall task. 
 
\end{abstract}

\section{Introduction} 
    Large Language Models (LLMs) and Chain-of-Thought techniques (CoT) continue to demonstrate impressive performance across a wide range of Natural Language Processing (NLP) tasks~\citep{singh2025openai,adcock2026llama,llama3,yang2025qwen3,NEURIPS2020_1457c0d6,NEURIPS2022_9d560961}. With the rapid advancement of LLMs, logical reasoning has become a crucial research topic, particularly in the context of explainable Artificial Intelligence (AI). Recent works show that LLMs still struggle with complex reasoning tasks~\citep{huang-chang-2023-towards,yee2024faithful,ranaldi-etal-2025-improving,ijcai2025p1155}. Numerous approaches have been proposed to improve the reasoning capabilities of LLMs, including prompt engineering~\citep{xu-etal-2024-faithful,nguyen2025noniterative,ranaldi-etal-2025-improving}, fine-tuning~\citep{feng-etal-2024-language,xu-etal-2024-symbol}, and the use of external symbolic solvers~\citep{ye2023satlm,pan-etal-2023-logic,xu-etal-2024-symbol}. 
    In contrast to previous works, this study does not focus on developing a sophisticated state-of-the-art reasoning framework. Instead, it investigates a fundamental research question: \textit{(RQ) What mechanisms do LLMs internally employ to solve logical reasoning tasks?}
    
\begin{figure*}[!htbp]
        \centering 
        \includegraphics[width=\linewidth, keepaspectratio, 
                trim={0.5cm 0 0.5cm 0cm }, page=1, clip=true]{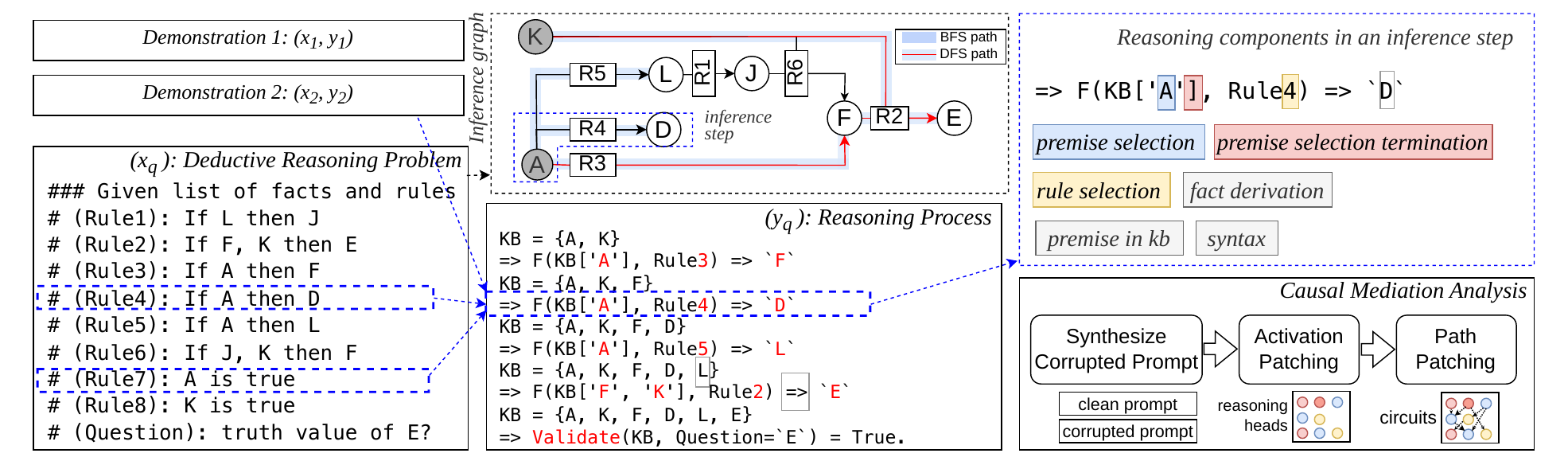} 
        \includegraphics[width=\linewidth, keepaspectratio, 
                trim={0.5cm 0.3cm 0cm 0cm }, page=1, clip=true]{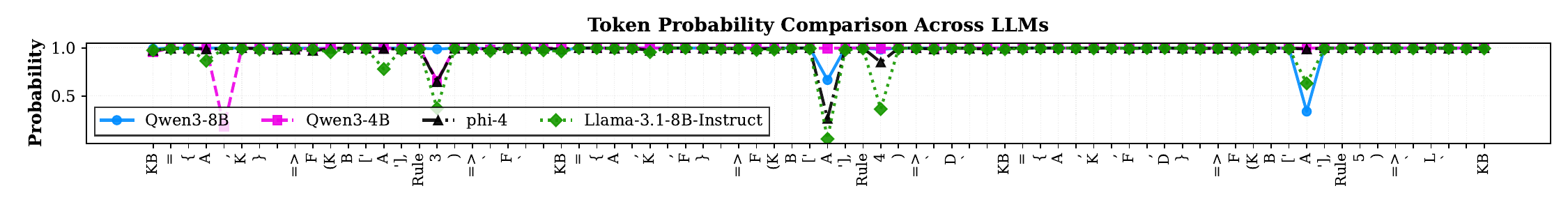} 
        \caption{Overview of this framework. Given a deductive reasoning problem description ($x_q$) together with a few-shot prompt containing demonstrations $[(x_1, y_1), (x_2, y_2), x_q]$ (left side of the figure), the LLM is expected to generate a Chain-of-Thought reasoning process that follows the demonstrated template. The deductive reasoning problem can be represented as an inference graph composed of multiple inference steps, each of which can be further decomposed into salient reasoning components (e.g., \textit{premise selection}) that are indicated by low-probability tokens across diverse LLMs (bottom graph in this figure). Subsequently, Causal Mediation Analysis techniques are employed to investigate the reasoning circuits of LLMs.
        }\label{fig_overview}
    \end{figure*}

    %
    Deductive reasoning in realistic settings remains a challenging task, as it requires multi-hop reasoning to determine whether a conclusion logically follows from a given set of premises \citep{tafjord-etal-2021-proofwriter,sun-etal-2024-determlr}.  
    In addition, recent studies \citep{nguyen2025noniterative,ranaldi-etal-2025-improving,xu-etal-2024-faithful} have shown that incorporating symbolic expressions into CoT prompting can improve the faithfulness of the LLM reasoning process. Therefore, we adopt the \textit{Symbolic-Aided CoT} prompting format proposed by \citet{nguyen2025noniterative} to effectively formalize the entire problem.
    In detail, we model the entire reasoning process as an inference graph constructed over predefined facts and rules (an example of a deductive reasoning problem and its corresponding inference graph in Figure~\ref{fig_overview}).
    Under this formulation, the reasoning problem can be viewed as a graph traversal process, which can be employed to identify a valid reasoning path from the initial facts to the target node representing the query. 
    %
    In this work, to address the aforementioned research question, we hypothesize that \textit{LLMs can ``learn'', at an abstract level, the meaning of each inference step in the reasoning process through few-shot demonstrations}. Here, ``learn'' refers to the phenomenon in which LLMs activate a specific subset of attention heads that serve as the primary mediators of the causal effect underlying the execution of inference steps and adherence to the designed graph-traversal algorithm.

    We adopt causal mediation analysis (CMA) \citep{10.5555/2074022.2074073,NEURIPS2020_92650b2e} techniques within the field of Mechanistic Interpretability to analyze the internal components responsible for the logical reasoning process and the interactions among them, commonly referred to as a \textit{circuit}  \citep{olah2020zoom,meng2022locating,wang2023interpretability}. 
    Diverse circuits in LLMs have been explored for specific tasks such as 
    arithmetic reasoning \citep{stolfo-etal-2023-mechanistic}, syllogistic reasoning \citep{kim-etal-2025-reasoning}, and propositional logical reasoning \citep{hong2026a}. 
    However, \textit{previous works have primarily focused on simple input–output settings, lacking the complexity of real-world scenarios in which the output includes a multi-step reasoning process preceding the final answer}. Our work investigates the circuits responsible for LLMs executing a reasoning strategy across multiple inference steps in CoT output. 
    % ...

    Overall, we first conduct preliminary experiments based on synthesized data to analyze the token positions that challenge LLMs in decoding the gold reasoning (referred to as \textit{uncertain tokens}), and then apply CMA techniques to inspect which internal components affect logit changes at these uncertain tokens (bottom graph in Figure~\ref{fig_overview}). We observe that high-confidence tokens are primarily associated with syntax or unambiguous reasoning actions. In contrast, uncertain tokens are the initial tokens that play a crucial role in steering the reasoning process. Decoding these tokens is equivalent to satisfying multiple implicit constraints.
    
    For example, on the top-right side of Figure~\ref{fig_overview}, consider the position of the \textit{premise selection} token, ``\texttt{A}'' (blue bounding rectangle). Three constraints must be satisfied: (1) the selected premise must correspond to a fact that has already been proven true and appears in the \texttt{KB} snapshot variable; (2) the selected premise must satisfy at least one applicable rule (in this case, \texttt{Rule3}, \texttt{Rule4}, and \texttt{Rule5} are valid); and (3) the selected premise must follow the traversal algorithm implied in the demonstrations (the breadth-first search (BFS) algorithm prefers selecting premise ``\texttt{A}'', whereas the depth-first search (DFS) algorithm prefers selecting premise ``\texttt{F}''). Similarly, the \textit{premise selection termination} token (red bounding rectangle) determines whether the model should continue selecting additional premises for the current inference step. The \textit{rule selection} token (yellow bounding rectangle) then determines which rule should be applied, given the selected premises.
    
    Interestingly, these uncertain positions remain challenging across diverse LLM families and model sizes.
    Further details of the preliminary experimental setup and statistical results are provided in Section~\ref{sec_prilim_exp}. Based on the preliminary experimental results, we categorize three major reasoning components within an inference step: \textit{premise selection}, \textit{premise selection termination}, \textit{rule selection}.
    Next, we leverage CMA techniques to uncover the \textit{circuit network}, a collection of circuits where each circuit handles one reasoning component and exhibits strong interactions with other circuits in the network. 
    We synthesize a new deductive logical reasoning dataset containing pairs of clean and corrupted prompts for each reasoning component. By applying activation patching and path patching techniques \citep{meng2022locating,wang2023interpretability}, we extract the important subset of attention heads responsible for each reasoning component, as well as the graph of information flow transferred among them. We further validate the generalization of these circuits by knocking them out of LLMs and evaluating the modified models on well-known deductive logical reasoning benchmarks, including ProntoQA \cite{saparov2023language}, ProofWriter \citep{tafjord-etal-2021-proofwriter}, and the general knowledge benchmark MMLU \cite{hendrycks2021measuring}. The performance on deductive logical reasoning tasks decreases dramatically, while performance on general knowledge tasks exhibits only a slight decrease, thereby confirming the importance and generality of our discovered circuits for deductive logical reasoning.

    

\section{Preliminary Experiment\label{sec_prilim_exp}}
    In this experiment, we aim to identify which tokens are \textit{important} for steering the deductive logical reasoning behavior of LLMs within the CoT reasoning process. As illustrated by the token probability scores in Figure~\ref{fig_overview}, these uncertain tokens are important because, when they are correctly predicted, the resulting logical reasoning process produces the correct final answer to the question.
    
    \paragraph{Dataset Construction.} We adopt the \textit{Symbolic-aided CoT} prompting format \citep{nguyen2025noniterative} and construct a synthesized dataset, where: (1) the premises are simplified and randomly selected from the uppercase alphabet characters, \texttt{A}–\texttt{Z}, following prior work \cite{hong2026a,kim-etal-2025-reasoning}; (2) each sample contains $k$ demonstrations; (3) the total number of rules and facts is randomly generated between 8 and 18; and (4) the BFS algorithm is used to generate the reasoning chains. In addition, we filter out all ambiguous samples in which the question cannot be logically inferred from the given rules and facts. Ultimately, we obtain a synthesized dataset, denoted as $\mathcal{D}^{\mathrm{syn}}_{k\,\mathrm{shots}}$ with 500 samples for each $k \in \{2, 3, 5, 7, 9\}$.
    
    \paragraph{Experimental Setup.} We select 10\% of $\mathcal{D}^{\mathrm{syn}}_{9\mathrm{shots}}$ and feed all data points into four LLMs (\texttt{Llama-3.1-8B-Instruct}, \texttt{Qwen3-8B}, \texttt{Phi-4}, and \texttt{Qwen3-4B} supported by \textit{transformers} library), while caching the token probabilities from the last five shots.
    In addition, we categorize each token in the reasoning chain according to its role within each inference step of the overall reasoning process. Specifically, we define six token types (corresponding to six reasoning components), as shown in Table~\ref{tab_token_roles_tagging}. Tokens with probabilities lower than 0.8 are identified as uncertain tokens and are subsequently used for analysis.
    \begin{table}[!htbp]
        \centering
        \caption{Reasoning component categories. The reasoning problem example is the same as Figure~\ref{fig_overview}. 
        \label{tab_token_roles_tagging} }
        \resizebox{\linewidth}{!}{%
            \begin{tabular}{p{0.67\linewidth}p{0.65\linewidth}}
                \toprule    
                    \textbf{Example}& \textbf{Colored Reasoning Component} \\\midrule
                    \texttt{KB = \{\textcolor{orange}{A}, \textcolor{orange}{K}\textcolor{black}{,} \textcolor{orange}{F}\textcolor{black}{\}}} &  syntax,  \textcolor{orange}{premise in KB},\\
                    
                    \texttt{{=> F(KB['}\textcolor{blue}{A}'\textcolor{red}{]}, Rule\textcolor{olive}{4}) => `\textcolor{OliveGreen}{D}` } \newline 
                    \texttt{{=> F(KB['}\textcolor{blue}{F}'\textcolor{red}{,} '\textcolor{blue}{K}'] Rule\textcolor{olive}{2}) => `\textcolor{OliveGreen}{E}` } 
                    &  \textcolor{blue}{premise selection}, \textcolor{red}{premise selection termination}, \textcolor{olive}{rule selection}, \textcolor{OliveGreen}{fact derivation}
                \\  
                  \bottomrule
            \end{tabular} 
        } 
    \end{table}
        
    \paragraph{Result Analysis and Motivation.}  
       We report the percentage of uncertain reasoning components and the distribution of their probabilities for \texttt{Llama-3.1-8B-Instruct}, \texttt{Qwen3-8B}, \texttt{Phi-4}, and \texttt{Qwen3-4B} in Figures~\ref{fig_low_confidence_tok_apdx} and~\ref{fig_low_confidence_tok} in Appendix~\ref{apd_prelim}.

        Our analysis reveals that three reasoning components account for the majority of the uncertainty, namely:   $\mathcal{R} = \{\mathrm{premise\,\, selection}, $ $ \mathrm{premise\,\, selection\,\, termination}, $ $ \mathrm{rule\,\, selection}\}$. As discussed in the introduction, these components play a critical role in guiding the reasoning process. In contrast, most other tokens in the reasoning chain are assigned high probabilities (i.e., certain tokens). Therefore, if LLMs can correctly decode these uncertain tokens, they are likely to generate gold reasoning paths. This observation strongly motivates the investigation of the internal mechanisms or circuits responsible for these reasoning components.
        
        Although the synthesized dataset does not capture certain real-world complexities, such as variations in premise complexity, it preserves the intrinsic challenges of logical reasoning through multi-hop reasoning and the number of rules that must be processed. In practice, on this synthesized dataset, \texttt{Qwen3-8B} and \texttt{Llama-3.1-8B-Instruct} achieve inference-step accuracies of approximately\footnote{Detailed experimental results are provided in Section~\ref{sec_main_experimental_results}.} 65\% and 50\%, respectively, highlighting the difficulty of the task. These findings further support the necessity of our research. We argue that the roles of attention heads in LLMs are shared across datasets that use the same task and format. To this end, we also conduct experiments on two well-known logical reasoning datasets, ProofWriter and ProntoQA, to further validate the generalization capability of our approach.
        

\section{Methodology}
In the main target, we performed CMA techniques   \citep{olah2020zoom,meng2022locating,wang2023interpretability,todd2024function} to discover the important heads and circuits in transformer decoder-only LLMs \cite{NIPS2017_3f5ee243} for each reasoning component ($r$) in $\mathcal{R}$. 

\subsection{Background and Notation}
Given a transformer-based LLM denoted by $f$ and a $k$-shot prompt 
$p = [(x_1, y_1), \ldots, (x_k, y_k), x_q]$, the model sequentially predicts tokens to generate the output string $y_q = f(p)$. At the output layer, $y_q$ is decoded as a sequence of probability distributions over the vocabulary $\mathcal{V}$, represented as $\mathbb{P} \in [0,1]^{|y_q| \times |\mathcal{V}|}$.
For simplification in this paper, when analyzing a specific \textit{reasoning token} $t^r$ at a known position $pos(t^r)$ in $y_q$, we use the notation $f(p)[t^r] = \mathbb{P}_{pos(t^r)}[t^r] \in [0,1]$ to denote the probability assigned to token $t^r$ at its position. For example, given ``\texttt{A}'' at position 13 in the output for $r = \mathrm{premise\,\, selection}$, then $f(p)[\texttt{A}]$ represents the probability $\mathbb{P}_{13}[\texttt{A}]$.

The residual stream of model $f$ at layer $\ell$ is computed as the sum of the previous layer's hidden state, the projected multi-layer perceptron (MLP) output ($m_{\ell}$), and the projected attention outputs $a_{\ell,j}$, defined as:
$h_{\ell} = h_{\ell - 1} + m_{\ell} + \sum_{j} a_{\ell j} \label{eq_transformer_base}$
where $j < J$ and $\ell < L$, with $L$ and $J$ denoting the number of layers and attention heads in $f$, respectively. For the first layer ($\ell = 0$), $h_{\ell-1}$ is initialized from the embedding layer.

Prior work in mechanistic interpretability (MI) has established that attention layers primarily perform information routing and relational composition, while MLP layers function as key-value memory stores for factual knowledge \citep{elhage2021mathematical,meng2022locating}, as further supported by the observations of \citet{hong2026a}. Therefore, in this work, we focus only on the attention components, $a_{\ell j}$.

\subsection{Circuit Discovery}
    Given a reasoning component $r \in \mathcal{R}$, we aim to discover the mechanism of model $f$ for decoding this component.  
    Firstly, we adopt the activation patching technique \cite{NEURIPS2020_92650b2e,todd2024function} to score the contribution of each attention head. Then, we utilize the path patching \cite{wang2023interpretability} for computing the amount of information transferred between the pair of heads. 
    
    \paragraph{Data preparation.} 
    %
    \begin{table*}[!htbp]
        \small
        \centering
        \caption{ Procedure to synthesize the pair of clean and corrupted prompts. The corrupted position is the causal elements that affect the reasoning component of interest.
        \label{tab_synthesize_data_cc} }
        \resizebox{\linewidth}{!}{%
            \begin{tabular}{p{0.11\linewidth}p{0.10\linewidth}p{0.4\linewidth}p{0.44\linewidth}}
                \toprule    
                    \textbf{Corrupted Position} &\textbf{Component of Interest}& \textbf{Procedure to Corrupt Prompt} & \textbf{Target} \textrightarrow\, \texttt{[Head roles]}  \\\midrule 
                Facts\newline\textit{(example in~Table~\ref{tab_example_corrupt_PS}}) & \texttt{premise selection}  &  (Corrupt type: $c=1$) Step 1. Randomly modify a used fact. \textrightarrow\, Step 2. Replace the corresponding premise change in the KB snapshot. & 
                
                Localize which heads \textit{read the facts} by patching the causal span (corrupted positions), and which heads handle \textit{premise selection based on the facts} by patching the previous (preceding) token of the component of interest. \textrightarrow\, \texttt{[Read Fact, Select Premise]} \\ \midrule
                Rules \newline \textit{(example in~Table~\ref{tab_example_corrupt_PST}}) &  \texttt{premise selection termination}\newline &  ($c=2$) Step 1. Choose the last used rule in the reasoning chain if it has a single premise condition, \texttt{Rx}  \textrightarrow\, Step 2. Replace the condition in \texttt{Rx} with an unseen premise. 
                    \textrightarrow\, Step 3. Randomly choose an unused rule having a double-premise condition, \texttt{Ry}  \textrightarrow\, Step 4. Replace the two conditional premises in \texttt{Ry} with: one conditional premise from \texttt{Rx} and one proven premise. & 
                    Localize which heads \textit{read the rule condition} by patching the causal span (both rules \texttt{Rx,Ry}), and which heads handle the \textit{matching of rule conditions with proven premises} by patching the previous token of the component of interest. \textrightarrow\, \texttt{[Read Rule Condition, Match Rule Condition]}
                \\\midrule
                Rules  \newline \textit{(example in~Table~\ref{tab_example_corrupt_RS}})  & \texttt{rule selection}  &  ($c=3$) Step 1. Randomly choose one used rule in the reasoning chain, \texttt{Rx} \textrightarrow\, Step 2. Replace the conditional premise of \texttt{Rx} with an unseen premise. & 
                    Localize which heads \textit{read the rule content} by patching the causal span (content of \texttt{Rx}) and which heads handle \textit{rule selection based on premise matching} by patching the preceding token.  \textrightarrow\, \texttt{[Read Rule, Select Rule]} \\ \midrule
                Demonstrations  \newline\textit{(example in~Table~\ref{tab_example_corrupt_alg}}) & \texttt{premise selection}   & ($c=4$) Step 1. Use a different traversal algorithm in demonstrations, ensuring that the corrupted reasoning chain is length equivalent to the clean reasoning chain for each demonstration.   & 
                    Localize which heads \textit{read} and \textit{implement the strategy of the traversal algorithm} from few-shot demonstrations by patching \textit{demonstrations} with the previous token of the component of interest. \textrightarrow\, \texttt{[Read Traversal Alg., Implement Traversal Alg.]}\\  
                  \bottomrule
            \end{tabular} 
    
        } 
    \end{table*}
    Activation patching, which reveals circuit components by measuring how restoring clean activations recovers model performance on corrupted inputs,  requires counterfactual pairs that differ precisely at the component of interest  \cite{NEURIPS2020_92650b2e,todd2024function}. Therefore, for each component, we construct a synthesized dataset ($\mathcal{D}^\mathrm{pair}$) where clean  ($p$) and corrupted prompts ($\tilde{p}$) share identical structure but differ in their causal context (Algorithm~\ref{alg:corrupt_generation} in Appendix~\ref{apd_alg_syn_data}). In these datasets, corrupted prompts systematically modify \textit{causal} elements (e.g., fact values or rule definitions) to induce a different token choice at the reasoning component position (e.g., \textit{premise selection}), creating a controlled comparison for activation patching.
    Table~\ref{tab_synthesize_data_cc} shows the detailed comparison of our synthesized procedure ($\textsc{Corrupt}$ procedure in Algorithm~\ref{alg:corrupt_generation}). 

    \paragraph{Activation Patching.}  
    Given the clean-corrupted prompting data pairs generated for corruption type $c$ (denoted by $\mathcal{D}^{c}$ for simplicity) from the previous step, the roles of LLM attention heads can be identified by measuring the average indirect effect (AIE) at the reasoning component of interest ($t^r$). The AIE is computed by patching clean activations into the corrupted forward pass, quantifying each head's contribution to recovering the model's clean-run behavior at position $t^r$.
    \begin{equation} \small
        \mathrm{AIE}_{\ell j}   = \frac{1}{|\mathcal{D}^{c}|} 
        \sum_{\tilde{p}_i \in  \mathcal{D}^{c}}   \big(
        f(\tilde{p}_i \mid a_{\ell j \mathbf{s}} =  a_{\ell j \mathbf{s}}(p_i))[t_i^r] 
        - f(\tilde{p})[t_i^r]\big)
    \end{equation}
    where 
    $a_{\ell j \mathbf{s}}(p_i)$ denotes the activation of head $j$ in layer $\ell$ at causal positions ($\mathbf{s}$ -- corrupted positions cached during the data synthesizing process) when the clean prompt is fed through model $f$. 

    Notably, we interpret the \textit{causal reading information heads} (e.g., \texttt{Read Fact}) when the patching positions correspond to the causal spans (corrupted positions). In contrast, the \textit{reasoning component decision heads} (e.g., \texttt{Select Premise}) are interpreted when the patching positions correspond to the \textit{preceding token} of the focal reasoning component. We then select the top-$k$ attention heads  ($\mathcal{A}^c$) with the highest causal indirect effect scores, which are considered responsible for corruption type $c$ within the reasoning component of interest.

    \paragraph{Path Patching.}   
    Path patching works by systematically corrupting and restoring activations between pairs of components to measure their causal influence on model outputs.  
    Given the set of important attention heads $\mathcal{A} = \{\mathcal{A}^c\}$, we aim to measure which heads communicate which information and quantify the strength of information flow. The combination of all subsets of heads responsible for each reasoning component reveals the interactions among reasoning circuits.
    For each pair of heads $(\mathrm{head}_{\mathrm{emit}}, \mathrm{head}_{\mathrm{rec}}) = ((\ell_1, j_1), (\ell_2, j_2)) \in \mathcal{A}\times \mathcal{A}$ where ($\ell_1 < \ell_2$), following \citet{wang2023interpretability}, we first perform a forward pass on the clean prompt while corrupting the activation at $\mathrm{head}_{\mathrm{emit}}$, which in turn corrupts the causally dependent activation at $\mathrm{head}_{\mathrm{rec}}$.  We then patch only this corrupted activation of $\mathrm{head}_{\mathrm{rec}}$ into a second forward pass of the clean prompt, allowing us to isolate and measure the causal dependence between this pair of heads. 
    
    {\small
    \begin{align}  
        \tilde{a}_{\ell_2 j_2 \mathbf{s}}  =& {a}_{\ell_2 j_2} ({p}_i \mid a_{\ell_1 j_1  \mathbf{s}} =  {a}_{\ell_1 j_1 \mathbf{s}}(\tilde{p}_i)) 
        \\
        \mathrm{S}_{i,(\ell_1 j_1)\rightarrow(\ell_2 j_2)}   =&  
        f({p}_i \mid a_{\ell_2 j_2 \mathbf{s}} =  \tilde{a}_{\ell_2 j_2 \mathbf{s}})[t_i^r] - f({p_i})[t_i^r] \notag \\ 
        \mathrm{S}_{(\ell_1 j_1)\rightarrow(\ell_2 j_2)}   =& \frac{1}{|\mathcal{D}^{c}|} 
        \sum_{p_i \in  \mathcal{D}^{c}}   \big( \mathrm{S}_{i,(\ell_1 j_1)\rightarrow(\ell_2 j_2)}\big)
    \end{align}
    } 
    
    {\noindent}Finally, we obtained $\mathrm{S}_{(\ell_1 j_1)\rightarrow(\ell_2 j_2)}$, a ranking score that measures the average causal effect between pairs of attention heads in the reasoning-responsible subset of heads, $\mathcal{A}$. 
    \begin{figure*}[!htbp]
        \centering  
        \includegraphics[width=1.\linewidth, keepaspectratio, 
                trim={ 0cm 0 8.5cm 0}, page=1, clip=true]{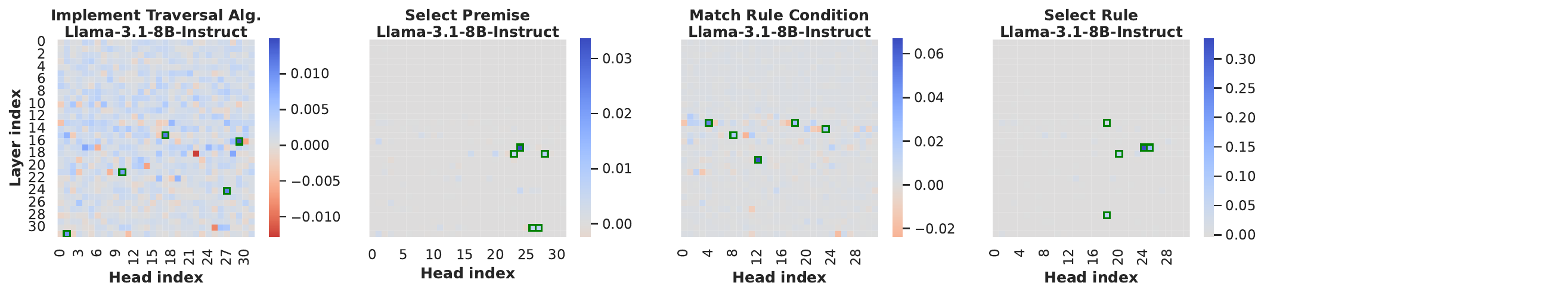}  
        \includegraphics[width=1\linewidth, keepaspectratio, 
                trim={0.15cm 0.45cm 7.33cm 0.89cm}, page=1, clip=true]{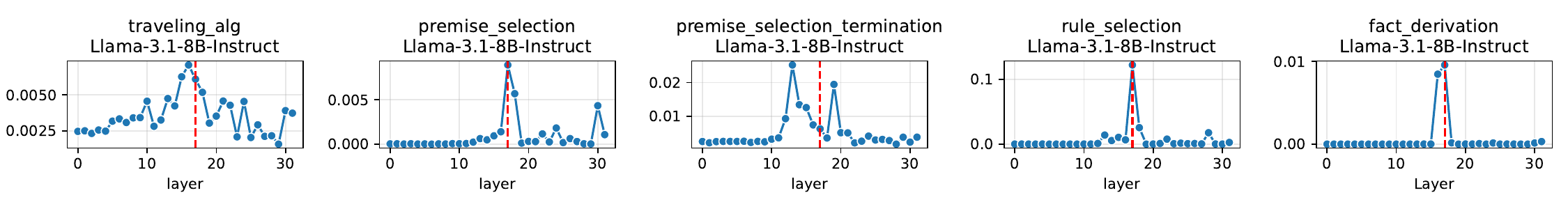}
    \caption{Distribution of AIE scores across layers and attention heads on \texttt{Llama-3.1-8B-Instruct} model. In the heatmap, the five highest-scoring heads are highlighted with green rectangles. In the line chart, the head role is aligned with the heat map; each layer score is computed by averaging the AIE scores of the top 15\% highest-scoring heads in that layer; the red dashed line shows the layer most responsible for \textit{rule selection}.}\label{fig_llama3_8b__aie_distdist}
    \end{figure*}

\section{Experimental Results \label{sec_main_experimental_results}}
    In this section, we present the experimental results and analyze the mechanism by which LLMs handle deductive logical reasoning tasks.

\subsection{Localizing Attention Heads} 
    The distribution of AIE scores across layers and attention heads in the \texttt{Llama-3.1-8B-Instruct} model is shown in Figure~\ref{fig_llama3_8b__aie_distdist} (the full version is provided in Figure~\ref{fig_apd_llama3_8b__aie_distdist}). Results for the other models are presented in Figures~\ref{fig_Qwen3_8B__aie_distdist},\ref{fig_Qwen3_4B__aie_distdist}, and \ref{fig_phi_4__aie_distdist}. Additionally, the layer score for each reasoning decision head role (line chart) is computed by averaging the AIE scores of the top 15\% highest-scoring heads within each layer.

\paragraph{Reading Head.} 
    The results in Figure~\ref{fig_apd_llama3_8b__aie_distdist} reveal that causal information-reading heads (e.g., \texttt{Read Fact}) are concentrated in earlier layers compared to decision-making heads (e.g., \texttt{Select Premise}). This trend is consistent across all evaluated LLMs, demonstrating the generalizability of this observation. This distribution is intuitive: information-reading heads extract relevant facts from the problem description and propagate them through the residual stream to higher layers, where subsequent logical reasoning operations are performed.

\paragraph{Decision Head.}
    Comparing reasoning decision heads across LLMs (heatmaps in Figures~\ref{fig_llama3_8b__aie_distdist},\ref{fig_Qwen3_8B__aie_distdist},\ref{fig_Qwen3_4B__aie_distdist}, and~\ref{fig_phi_4__aie_distdist}), we observe that heads responsible for \textit{matching rule conditions} are concentrated in the middle layers and correspond to \textit{the earliest reasoning stage} in the computational pipeline. This observation is intuitively consistent with human reasoning, where rule conditions must first be validated before determining which rules should be applied.
    In addition, heads responsible for \textit{rule selection} exhibit high sparsity, with a small number of heads accounting for a dominant proportion of the effect. Notably, the highest AIE score in \texttt{Llama-3.1-8B-Instruct} exceeds 30\%, suggesting that the model relies heavily on a single head for rule selection decisions. A similar pattern is observed in the \texttt{Qwen} models, where the peak AIE scores exceed 12\%. We hypothesize that this sparsity arises because rule selection occurs after premise selection in the reasoning pipeline of prompting design (Figure~\ref{fig_overview}), making it a more deterministic operation that can be handled by a small number of specialized, high-impact heads.
    
    On the other hand, the line graphs on Figure~\ref{fig_llama3_8b__aie_distdist} reveal a consistent \textit{temporal computational structure} across all LLMs: matching rule condition \textrightarrow{} implementing traversal algorithm \textrightarrow{} selecting premise and rule \textrightarrow{} premise decision (termination).
    \begin{figure}[!htbp]
            \centering  
            \includegraphics[width=0.99\linewidth, keepaspectratio, 
                    trim={ 3cm 2.cm 5cm 1.2cm }, page=1, clip=true]{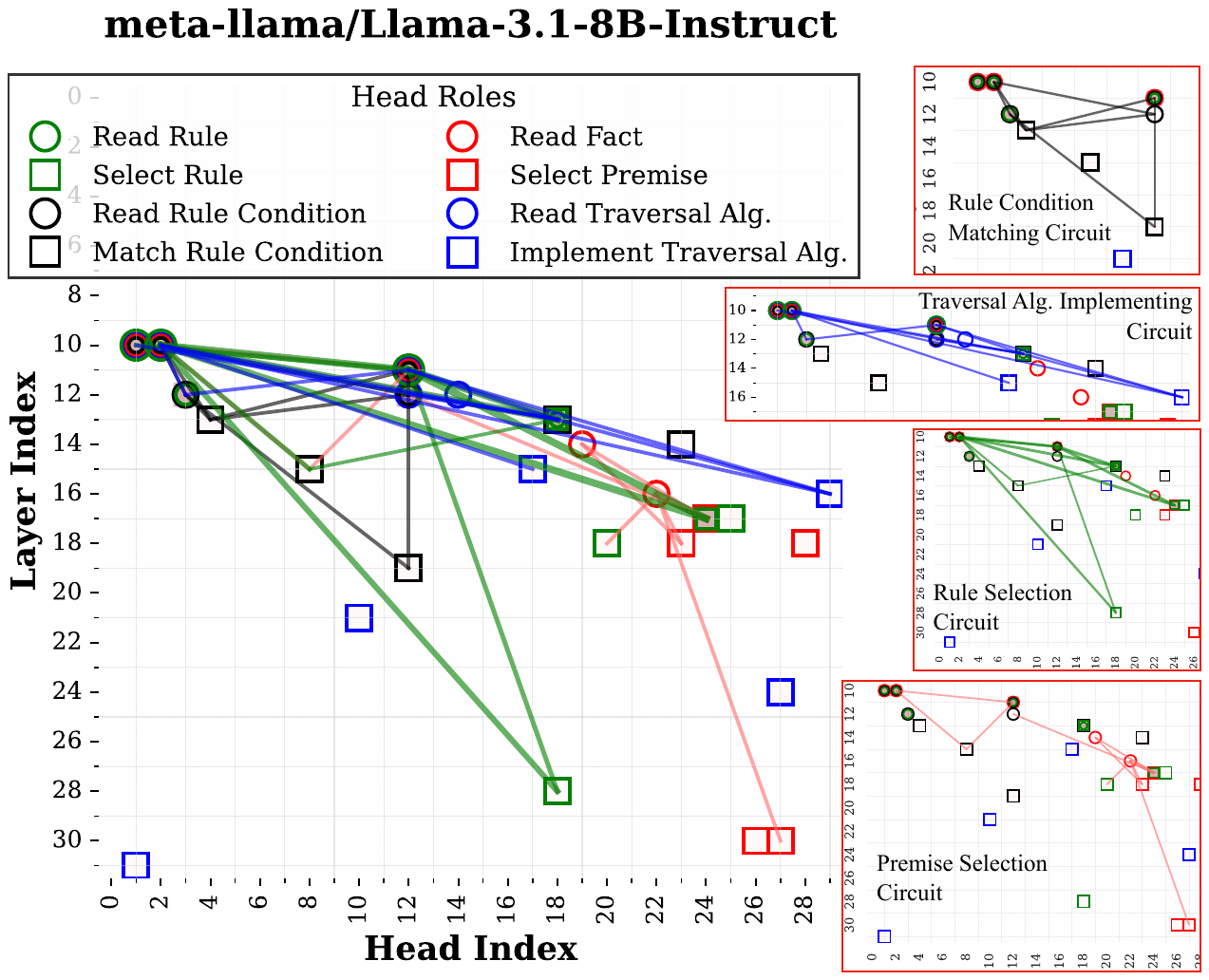} 
            \caption{\textit{Circuit network} on top-5 attention head scores associated with reasoning components.}\label{fig_llama_headroles}
        \end{figure}
\subsection{Circuit Network}   
    
    Here, we analyze a collection of sub-circuits corresponding to reasoning components, referred to as a \textit{circuit network}. For each reasoning component, we analyze the top-5 heads for each reasoning role and the top-10 strongest causal effects between pairs of heads computed by path patching score ($\mathrm{S}$)  (Figure~\ref{fig_llama_headroles},~\ref{fig_apd_llama_qw_8b_cir},~\ref{fig_apd_llama_qw_4b_cir},~\ref{fig_apd_phi4_cir}).
  
    \paragraph{Circuit Specialization.} The results show that for each circuit, each type of reasoning head typically transfers information corresponding to its role. For example, in the \textit{Rule Condition Matching} circuit, information about the rule condition is transferred from the reading heads to the decision heads (e.g., rule condition information is transferred from (L$11$H$12$ (stand for layer 11 head 12), L$12$H$3$) to L$19$H$12$, (L$10$H$2$, L$11$H$12$, L$12$H$12$) to L$11$H$4$). In addition, the information for each sub-reasoning task is transferred and integrated into the decision heads at the deeper layer (L$19$ for Rule Condition Matching, L$17$ and L$28$ for Rule Selection, L$18$ and L$30$ for Premise Selection). This clearly confirms the characteristic of  \textit{temporal computational structure} observed in the previous section.
    \paragraph{Circuit Interaction.}  Overall, many attention heads perform multiple reasoning sub-tasks (\textit{polysemantic attention head}) and the polysemantic decision heads are the center of integration of information. For instance,  L$10$H$1$, L$10$H$2$, and L$11$H$12$  perform three types of causal information reading: \textit{read rule condition}, \textit{read rule}, and \textit{read fact}; polysemantic decision head, L$17$H$24$, handles multiple reasoning decision sub-tasks. Different LLMs exhibit distinct polysemantic patterns: \texttt{Llama-3.1-8B-Instruct} shares heads primarily for causal reading roles, whereas \texttt{Qwen} models share heads across decision-making roles. 

\subsection{Ablation LR Heads Across Datasets} 
To further confirm the necessity of deductive LR heads, we evaluate their impact across our synthesized dataset $\mathcal{D}^{syn}$, presented in Section~\ref{sec_prilim_exp}, established LR benchmarks (ProntoQA \cite{saparov2023language}, ProofWriter \citep{tafjord-etal-2021-proofwriter}), and a general knowledge dataset (MMLU \cite{hendrycks2021measuring}). In both LR benchmarks, we apply \textit{Symbolic-Aided CoT} prompting following the setting from previous works \cite{nguyen2025noniterative,nguyen-etal-2026-improving}.
We ablate (knock out) the top-$k$ important heads by setting $a_{\ell j} = 0$ when computing $h_{\ell j}$ (Section~\ref{eq_transformer_base}) for all tokens associated with the ablated heads (both the prefill and decoding stage), for each reasoning component role identified by the AIE scores. We then measure the resulting performance degradation and compare it against random head ablation across LLMs.

\paragraph{Experimental Setup.} 
Since models differ in total head count, we set  $k=8$ for \texttt{phi-4} and $k=5$ for \texttt{Qwen} and \texttt{Llama} models. We evaluate six configurations: \textbf{\texttt{(baseline)}} evaluates original model performance; \textbf{\texttt{/Rand}} ablates 3\% of heads uniformly at random and average over three random independent run; \textbf{\texttt{/RS}} ablates {top-$k$} \textit{reading rule} and top-$k$ \textit{rule selection} heads ($\approx$ 1\% of total heads); \textbf{\texttt{/PS}} ablates top-$k$ \textit{reading fact} and top-$k$ \textit{premise selection} heads ($\approx$ 1\%); \textbf{\texttt{/PST}} ablates top-$k$ \textit{reading rule condition} and top-$k$ \textit{matching rule condition} heads ($\approx$ 1\%); \textbf{\texttt{/3Roles}} ablates all heads from the previous three configurations (\texttt{/RS}, \texttt{/PS}, and \texttt{/PST}, totaling $\approx$ 3\% of heads). 

For the evaluation, on MMLU, PronQA, and ProofWriter, we primarily used final-answer accuracy. For the synthesized data, we used inference-step accuracy, which measures the proportion of correct reasoning steps (or correct inference steps over the rule space) rather than considering only the final answer, which helps more accurately evaluate the reasoning ability of LLMs.
\begin{figure}[!htbp]
\centering  
\includegraphics[width=.99\linewidth, keepaspectratio, 
        trim={ 0.1cm 0.3cm 0.1cm 0.2cm}, page=1, clip=true]{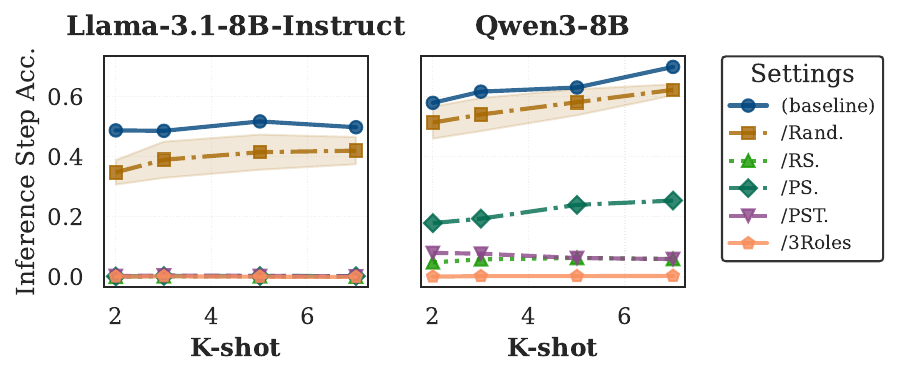} 
\caption{Inference step accuracy after knocking out top-$k$ logical reasoning heads on synthesized data.}\label{fig_knockout_syn_acc}
\end{figure}
\begin{figure}[!htbp]
\centering  
\includegraphics[width=.99\linewidth, keepaspectratio, 
        trim={ 0.3cm 0.2cm 0.1cm 0.2cm}, page=1, clip=true]{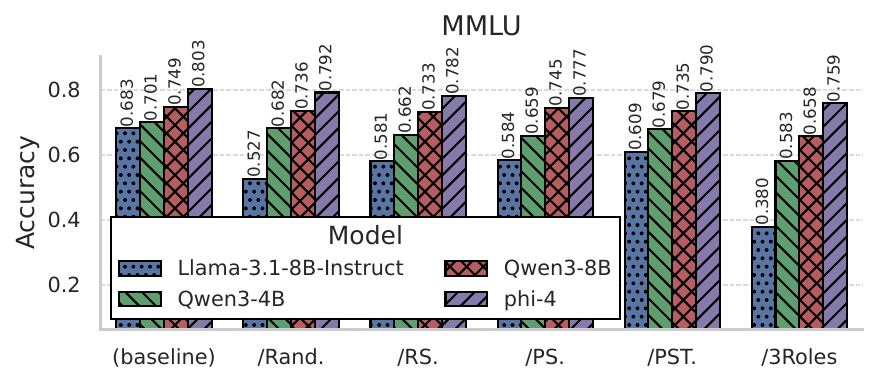}
\includegraphics[width=.99\linewidth, keepaspectratio, 
        trim={ 0.3cm 0.2cm 0.1cm 0.2cm}, page=1, clip=true]{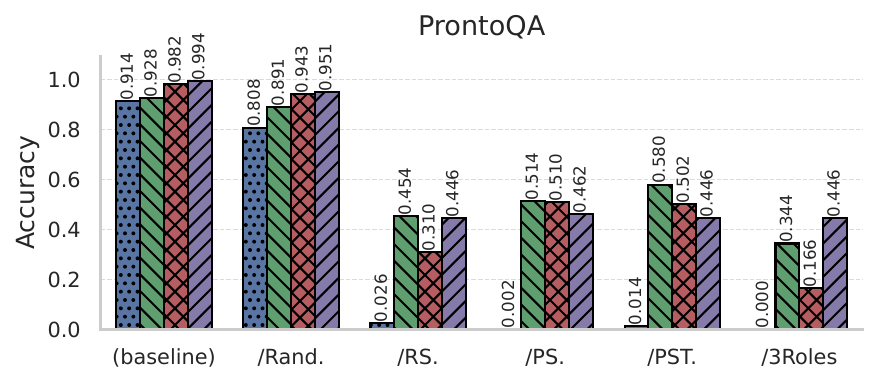}  
\caption{Effect of knocking out top-$k$ logical reasoning heads on MMLU and ProntoQA datasets.}\label{fig_lr_kb_knockout}
\end{figure}

\paragraph{Results and Analysis.} The results of this experiment are shown in Figures~\ref{fig_knockout_syn_acc},~\ref{fig_lr_kb_knockout} and in Figure~\ref{fig_apd_knockout_syn_acc},  Appendix~\ref{apd_knockout_head}.
On the synthesized data, ablating the LR heads significantly decreases performance across few-shot configurations compared to ablating uniformly random heads, even when the number of ablated heads is three times smaller in the \texttt{/RS}, \texttt{/PS}, and \texttt{/PST} settings. In particular, ablating the three important reasoning components (setting \texttt{/3Roles}) causes the reasoning ability of all LLMs to collapse to nearly zero.

On two well-known LR benchmarks, ProofWriter and ProntoQA, the results show the same trend across LLMs, demonstrating that the LR heads identified by our method on synthesized data generalize well and are not biased toward domain- or data-specific features.  In addition, we observe that the accuracies of 44.6\% on ProntoQA and 29.2\% on ProofWriter for \texttt{phi-4} (similar behavior observed for other models) are mainly due to random guessing of the final answer after generating an incorrect reasoning chain. This is because ProntoQA contains two possible answers (True/False), whereas ProofWriter contains three possible answers (True/False/Uncertain). In other words, the LLMs only preserve the reasoning format rather than performing a meaningful reasoning process.
On the MMLU dataset, ablating individual reasoning component types (\texttt{/RS}, \texttt{/PS}, \texttt{/PST}) yields performance drops comparable to random ablation. However, the \texttt{/3Roles} setting, which ablates all LR heads simultaneously, causes substantially larger degradation than random ablation of the same proportion. This suggests that while individual LR components have limited impact on general knowledge tasks, their collective contribution is significant, indicating that deductive reasoning heads are recruited even for broad knowledge retrieval and application.

\section{Related Work}
    \paragraph{Logical Reasoning in LLMs.} Recent studies reveal that LLMs struggle with complex reasoning tasks~\citep{yee2024faithful,ranaldi-etal-2025-improving,ijcai2025p1155}. To address these limitations, researchers have explored multiple approaches, including: prompt engineering~\citep{xu-etal-2024-faithful,nguyen2025noniterative,ranaldi-etal-2025-improving}, fine-tuning~\citep{feng-etal-2024-language,xu-etal-2024-symbol}, incorporating symbolic expressions into Chain-of-Thought prompting~\citep{nguyen2025noniterative,ranaldi-etal-2025-improving,xu-etal-2024-faithful}, leveraging external symbolic solvers where LLMs translate natural language into formal logic~\citep{pan-etal-2023-logic,ye2023satlm,olausson-etal-2023-linc,pmlr-v202-gao23f,yao2023react,xu-etal-2024-symbol}, and developing interactive prompting frameworks that decompose complex reasoning into manageable subtasks~\citep{zhou2023leasttomost,zhang2025cumulative,kazemi-etal-2023-lambada,feng-etal-2024-language,sun-etal-2024-determlr,lee-hwang-2025-symba}. Unlike these works, our work investigates deductive reasoning circuits in LLMs.

    \paragraph{Reasoning Circuits in LLMs.} 
    In recent works, various circuits have been discovered for specialized tasks, including factual knowledge recall \citep{meng2022locating}, indirect object identification \citep{stolfo-etal-2023-mechanistic}, time-specific information \citep{park-etal-2025-time}, arithmetic reasoning~\citep{stolfo-etal-2023-mechanistic}, abstract reasoning with emergent symbolic \cite{yang2025emergent}, syllogistic reasoning~\citep{kim-etal-2025-reasoning}, and propositional logical reasoning~\citep{hong2026a}. 
    However, prior circuit analysis work has focused on simple input-output mappings rather than multi-step reasoning processes. Our work addresses this gap by decomposing complex deductive reasoning into constituent reasoning components and identifying the circuits responsible for each step.


\section{Conclusion\label{sec_conclusion}}   
    In this work, we investigated the mechanisms underlying deductive reasoning in LLMs by localizing and characterizing the attention heads responsible for multi-step logical inference. Using causal mediation analysis, we identified specialized attention heads associated with different reasoning components. Our analysis reveals a hierarchical structure in which early-to-middle layers retrieve and process factual and rule-based information, while higher layers integrate this information and coordinate global reasoning strategies.
    We further found that reasoning-critical token positions consistently exhibit low confidence scores, reflecting the strict reasoning constraints induced by few-shot demonstrations. By ablating the identified reasoning heads across multiple datasets, including our synthesized benchmark, ProntoQA, ProofWriter, and MMLU, we showed that these circuits are essential for deductive reasoning, with their removal causing substantially larger performance degradation than random ablation.
    Overall, our findings provide mechanistic insights into how LLMs internalize abstract reasoning algorithms from limited demonstrations, revealing a sparse and modular circuit architecture that supports complex multi-hop inference. This work may contribute to improving reasoning robustness, interpretability, and targeted interventions for future language models.

\section*{Limitations} 
 Our analysis is primarily conducted on synthesized datasets with explicit logical structures, which may not fully generalize to free-form text reasoning or implicit chain-of-thought processes in naturalistic contexts. Additionally, we focus on attention mechanisms while other components (e.g., MLP layers) may also contribute to reasoning. The identified circuits are specific to the models and datasets studied; their generalization across different architectures, scales, and reasoning domains requires further investigation.




\bibliography{custom}

@article{llama3,
  publtype={informal},
  author={Abhimanyu Dubey and Abhinav Jauhri and Abhinav Pandey and Abhishek Kadian and Ahmad Al-Dahle and Aiesha Letman and Akhil Mathur and Alan Schelten and Amy Yang and Angela Fan and Anirudh Goyal and Anthony Hartshorn and Aobo Yang and Archi Mitra and Archie Sravankumar and Artem Korenev and Arthur Hinsvark and Arun Rao and Aston Zhang and Aurélien Rodriguez and Austen Gregerson and Ava Spataru and Baptiste Rozière and Bethany Biron and Binh Tang and Bobbie Chern and Charlotte Caucheteux and Chaya Nayak and Chloe Bi and Chris Marra and Chris McConnell and Christian Keller and Christophe Touret and Chunyang Wu and Corinne Wong and Cristian Canton Ferrer and Cyrus Nikolaidis and Damien Allonsius and Daniel Song and Danielle Pintz and Danny Livshits and David Esiobu and Dhruv Choudhary and Dhruv Mahajan and Diego Garcia-Olano and Diego Perino and Dieuwke Hupkes and Egor Lakomkin and Ehab AlBadawy and Elina Lobanova and Emily Dinan and Eric Michael Smith and Filip Radenovic and Frank Zhang and Gabriel Synnaeve and Gabrielle Lee and Georgia Lewis Anderson and Graeme Nail and Grégoire Mialon and Guan Pang and Guillem Cucurell and Hailey Nguyen and Hannah Korevaar and Hu Xu and Hugo Touvron and Iliyan Zarov and Imanol Arrieta Ibarra and Isabel M. Kloumann and Ishan Misra and Ivan Evtimov and Jade Copet and Jaewon Lee and Jan Geffert and Jana Vranes and Jason Park and Jay Mahadeokar and Jeet Shah and Jelmer van der Linde and Jennifer Billock and Jenny Hong and Jenya Lee and Jeremy Fu and Jianfeng Chi and Jianyu Huang and Jiawen Liu and Jie Wang and Jiecao Yu and Joanna Bitton and Joe Spisak and Jongsoo Park and Joseph Rocca and Joshua Johnstun and Joshua Saxe and Junteng Jia and Kalyan Vasuden Alwala and Kartikeya Upasani and Kate Plawiak and Ke Li and Kenneth Heafield and Kevin Stone and et al.},
  title={The Llama 3 Herd of Models},
  year={2024},
  cdate={1704067200000},
  journal={CoRR},
  volume={abs/2407.21783},
  url={https://doi.org/10.48550/arXiv.2407.21783}
}

@article{
zhang2025cumulative,
title={Cumulative Reasoning with Large Language Models},
author={Yifan Zhang and Jingqin Yang and Yang Yuan and Andrew C Yao},
journal={Transactions on Machine Learning Research},
issn={2835-8856},
year={2025},
url={https://openreview.net/forum?id=grW15p4eq2},
note={}
}

@inproceedings{
ye2023satlm,
title={Sat{LM}: Satisfiability-Aided Language Models Using Declarative Prompting},
author={Xi Ye and Qiaochu Chen and Isil Dillig and Greg Durrett},
booktitle={Thirty-seventh Conference on Neural Information Processing Systems},
year={2023},
url={https://openreview.net/forum?id=TqW5PL1Poi}
}

@InProceedings{pmlr-v202-gao23f,
  title = 	 {{PAL}: Program-aided Language Models},
  author =       {Gao, Luyu and Madaan, Aman and Zhou, Shuyan and Alon, Uri and Liu, Pengfei and Yang, Yiming and Callan, Jamie and Neubig, Graham},
  booktitle = 	 {Proceedings of the 40th International Conference on Machine Learning},
  pages = 	 {10764--10799},
  year = 	 {2023},
  editor = 	 {Krause, Andreas and Brunskill, Emma and Cho, Kyunghyun and Engelhardt, Barbara and Sabato, Sivan and Scarlett, Jonathan},
  volume = 	 {202},
  series = 	 {Proceedings of Machine Learning Research},
  month = 	 {23--29 Jul},
  publisher =    {PMLR},
  url = 	 {https://proceedings.mlr.press/v202/gao23f.html}
}

@inproceedings{pan-etal-2023-logic,
    title = "Logic-{LM}: Empowering Large Language Models with Symbolic Solvers for Faithful Logical Reasoning",
    author = "Pan, Liangming  and
      Albalak, Alon  and
      Wang, Xinyi  and
      Wang, William",
    editor = "Bouamor, Houda  and
      Pino, Juan  and
      Bali, Kalika",
    booktitle = "Findings of the Association for Computational Linguistics: EMNLP 2023",
    month = dec,
    year = "2023",
    address = "Singapore",
    publisher = "Association for Computational Linguistics",
    url = "https://aclanthology.org/2023.findings-emnlp.248/",
    doi = "10.18653/v1/2023.findings-emnlp.248",
    pages = "3806--3824"
}

@inproceedings{NEURIPS2020_1457c0d6,
 author = {Brown, Tom and Mann, Benjamin and Ryder, Nick and Subbiah, Melanie and Kaplan, Jared D and Dhariwal, Prafulla and Neelakantan, Arvind and Shyam, Pranav and Sastry, Girish and Askell, Amanda and Agarwal, Sandhini and Herbert-Voss, Ariel and Krueger, Gretchen and Henighan, Tom and Child, Rewon and Ramesh, Aditya and Ziegler, Daniel and Wu, Jeffrey and Winter, Clemens and Hesse, Chris and Chen, Mark and Sigler, Eric and Litwin, Mateusz and Gray, Scott and Chess, Benjamin and Clark, Jack and Berner, Christopher and McCandlish, Sam and Radford, Alec and Sutskever, Ilya and Amodei, Dario},
 booktitle = {Advances in Neural Information Processing Systems},
 editor = {H. Larochelle and M. Ranzato and R. Hadsell and M.F. Balcan and H. Lin},
 pages = {1877--1901},
 publisher = {Curran Associates, Inc.},
 title = {Language Models are Few-Shot Learners},
 url = {https://proceedings.neurips.cc/paper_files/paper/2020/file/1457c0d6bfcb4967418bfb8ac142f64a-Paper.pdf},
 volume = {33},
 year = {2020}
}

@inproceedings{xu-etal-2024-symbol,
    title = "Symbol-{LLM}: Towards Foundational Symbol-centric Interface For Large Language Models",
    author = "Xu, Fangzhi  and
      Wu, Zhiyong  and
      Sun, Qiushi  and
      Ren, Siyu  and
      Yuan, Fei  and
      Yuan, Shuai  and
      Lin, Qika  and
      Qiao, Yu  and
      Liu, Jun",
    editor = "Ku, Lun-Wei  and
      Martins, Andre  and
      Srikumar, Vivek",
    booktitle = "Proceedings of the 62nd Annual Meeting of the Association for Computational Linguistics (Volume 1: Long Papers)",
    month = aug,
    year = "2024",
    address = "Bangkok, Thailand",
    publisher = "Association for Computational Linguistics",
    url = "https://aclanthology.org/2024.acl-long.707/",
    doi = "10.18653/v1/2024.acl-long.707",
    pages = "13091--13116"
}

@inproceedings{
zhou2023leasttomost,
title={Least-to-Most Prompting Enables Complex Reasoning in Large Language Models},
author={Denny Zhou and Nathanael Sch{\"a}rli and Le Hou and Jason Wei and Nathan Scales and Xuezhi Wang and Dale Schuurmans and Claire Cui and Olivier Bousquet and Quoc V Le and Ed H. Chi},
booktitle={The Eleventh International Conference on Learning Representations },
year={2023},
url={https://openreview.net/forum?id=WZH7099tgfM}
}

@inproceedings{tafjord-etal-2021-proofwriter,
    title = "{P}roof{W}riter: Generating Implications, Proofs, and Abductive Statements over Natural Language",
    author = "Tafjord, Oyvind  and
      Dalvi, Bhavana  and
      Clark, Peter",
    editor = "Zong, Chengqing  and
      Xia, Fei  and
      Li, Wenjie  and
      Navigli, Roberto",
    booktitle = "Findings of the Association for Computational Linguistics: ACL-IJCNLP 2021",
    month = aug,
    year = "2021",
    address = "Online",
    publisher = "Association for Computational Linguistics",
    url = "https://aclanthology.org/2021.findings-acl.317/",
    doi = "10.18653/v1/2021.findings-acl.317",
    pages = "3621--3634"
}

@inproceedings{
yao2023react,
title={ReAct: Synergizing Reasoning and Acting in Language Models},
author={Shunyu Yao and Jeffrey Zhao and Dian Yu and Nan Du and Izhak Shafran and Karthik R Narasimhan and Yuan Cao},
booktitle={The Eleventh International Conference on Learning Representations },
year={2023},
url={https://openreview.net/forum?id=WE_vluYUL-X}
}

@inproceedings{
saparov2023language,
title={Language Models Are Greedy Reasoners: A Systematic Formal Analysis of Chain-of-Thought},
author={Abulhair Saparov and He He},
booktitle={The Eleventh International Conference on Learning Representations },
year={2023},
url={https://openreview.net/forum?id=qFVVBzXxR2V}
}

@article{yang2025qwen3,
  title={Qwen3 technical report},
  author={Yang, An and Li, Anfeng and Yang, Baosong and Zhang, Beichen and Hui, Binyuan and Zheng, Bo and Yu, Bowen and Gao, Chang and Huang, Chengen and Lv, Chenxu and others},
  journal={arXiv preprint arXiv:2505.09388},
  year={2025}
}

@inproceedings{feng-etal-2024-language,
    title = "Language Models can be Deductive Solvers",
    author = "Feng, Jiazhan  and
      Xu, Ruochen  and
      Hao, Junheng  and
      Sharma, Hiteshi  and
      Shen, Yelong  and
      Zhao, Dongyan  and
      Chen, Weizhu",
    editor = "Duh, Kevin  and
      Gomez, Helena  and
      Bethard, Steven",
    booktitle = "Findings of the Association for Computational Linguistics: NAACL 2024",
    month = jun,
    year = "2024",
    address = "Mexico City, Mexico",
    publisher = "Association for Computational Linguistics",
    url = "https://aclanthology.org/2024.findings-naacl.254/",
    doi = "10.18653/v1/2024.findings-naacl.254",
    pages = "4026--4042"
}

@inproceedings{sun-etal-2024-determlr,
    title = "{D}eterm{LR}: Augmenting {LLM}-based Logical Reasoning from Indeterminacy to Determinacy",
    author = "Sun, Hongda  and
      Xu, Weikai  and
      Liu, Wei  and
      Luan, Jian  and
      Wang, Bin  and
      Shang, Shuo  and
      Wen, Ji-Rong  and
      Yan, Rui",
    editor = "Ku, Lun-Wei  and
      Martins, Andre  and
      Srikumar, Vivek",
    booktitle = "Proceedings of the 62nd Annual Meeting of the Association for Computational Linguistics (Volume 1: Long Papers)",
    month = aug,
    year = "2024",
    address = "Bangkok, Thailand",
    publisher = "Association for Computational Linguistics",
    url = "https://aclanthology.org/2024.acl-long.531/",
    doi = "10.18653/v1/2024.acl-long.531",
    pages = "9828--9862"
}

@inproceedings{NEURIPS2022_9d560961,
 author = {Wei, Jason and Wang, Xuezhi and Schuurmans, Dale and Bosma, Maarten and ichter, brian and Xia, Fei and Chi, Ed and Le, Quoc V and Zhou, Denny},
 booktitle = {Advances in Neural Information Processing Systems},
 editor = {S. Koyejo and S. Mohamed and A. Agarwal and D. Belgrave and K. Cho and A. Oh},
 pages = {24824--24837},
 publisher = {Curran Associates, Inc.},
 title = {Chain-of-Thought Prompting Elicits Reasoning in Large Language Models},
 url = {https://proceedings.neurips.cc/paper_files/paper/2022/file/9d5609613524ecf4f15af0f7b31abca4-Paper-Conference.pdf},
 volume = {35},
 year = {2022}
}

@inproceedings{xu-etal-2024-faithful,
    title = "Faithful Logical Reasoning via Symbolic Chain-of-Thought",
    author = "Xu, Jundong  and
      Fei, Hao  and
      Pan, Liangming  and
      Liu, Qian  and
      Lee, Mong-Li  and
      Hsu, Wynne",
    editor = "Ku, Lun-Wei  and
      Martins, Andre  and
      Srikumar, Vivek",
    booktitle = "Proceedings of the 62nd Annual Meeting of the Association for Computational Linguistics (Volume 1: Long Papers)",
    month = aug,
    year = "2024",
    address = "Bangkok, Thailand",
    publisher = "Association for Computational Linguistics",
    url = "https://aclanthology.org/2024.acl-long.720/",
    doi = "10.18653/v1/2024.acl-long.720",
    pages = "13326--13365"
}

@inproceedings{NIPS2017_3f5ee243,
 author = {Vaswani, Ashish and Shazeer, Noam and Parmar, Niki and Uszkoreit, Jakob and Jones, Llion and Gomez, Aidan N and Kaiser, \L ukasz and Polosukhin, Illia},
 booktitle = {Advances in Neural Information Processing Systems},
 editor = {I. Guyon and U. Von Luxburg and S. Bengio and H. Wallach and R. Fergus and S. Vishwanathan and R. Garnett},
 pages = {},
 publisher = {Curran Associates, Inc.},
 title = {Attention is All you Need},
 url = {https://proceedings.neurips.cc/paper_files/paper/2017/file/3f5ee243547dee91fbd053c1c4a845aa-Paper.pdf},
 volume = {30},
 year = {2017}
}

@inproceedings{nguyen2025noniterative,
    title = "Non-Interactive Symbolic-Aided Chain-of-Thought for Logical Reasoning",
    author = "Nguyen, Minh-Phuong  and
      Dang, Tien  and
      Inoue, Naoya",
    editor = "Huang, Chu-Ren  and
      Harada, Yasunari  and
      Kim, Jong-Bok  and
      Huyen, Nguyen T.M.  and
      Huong, Le Thanh  and
      Hien, Pham  and
      Chersoni, Emmanuele  and
      Nguyen, Le Minh  and
      Roxas, Rachel Edita O{\~n}ate  and
      Dita, Sherly",
    booktitle = "Proceedings of the 39th Pacific {A}sia Conference on Language, Information and Computation",
    month = dec,
    year = "2025",
    address = "Hanoi, Vietnam",
    publisher = "Association for Computational Linguistics",
    url = "https://aclanthology.org/2025.paclic-1.29/",
    pages = "329--340"
}

@inproceedings{olausson-etal-2023-linc,
    title = "{LINC}: A Neurosymbolic Approach for Logical Reasoning by Combining Language Models with First-Order Logic Provers",
    author = "Olausson, Theo  and
      Gu, Alex  and
      Lipkin, Ben  and
      Zhang, Cedegao  and
      Solar-Lezama, Armando  and
      Tenenbaum, Joshua  and
      Levy, Roger",
    editor = "Bouamor, Houda  and
      Pino, Juan  and
      Bali, Kalika",
    booktitle = "Proceedings of the 2023 Conference on Empirical Methods in Natural Language Processing",
    month = dec,
    year = "2023",
    address = "Singapore",
    publisher = "Association for Computational Linguistics",
    url = "https://aclanthology.org/2023.emnlp-main.313/",
    doi = "10.18653/v1/2023.emnlp-main.313",
    pages = "5153--5176"
}

@inproceedings{kazemi-etal-2023-lambada,
    title = "{LAMBADA}: Backward Chaining for Automated Reasoning in Natural Language",
    author = "Kazemi, Mehran  and
      Kim, Najoung  and
      Bhatia, Deepti  and
      Xu, Xin  and
      Ramachandran, Deepak",
    editor = "Rogers, Anna  and
      Boyd-Graber, Jordan  and
      Okazaki, Naoaki",
    booktitle = "Proceedings of the 61st Annual Meeting of the Association for Computational Linguistics (Volume 1: Long Papers)",
    month = jul,
    year = "2023",
    address = "Toronto, Canada",
    publisher = "Association for Computational Linguistics",
    url = "https://aclanthology.org/2023.acl-long.361/",
    doi = "10.18653/v1/2023.acl-long.361",
    pages = "6547--6568"
}

@inproceedings{lee-hwang-2025-symba,
    title = "{S}ym{B}a: Symbolic Backward Chaining for Structured Natural Language Reasoning",
    author = "Lee, Jinu  and
      Hwang, Wonseok",
    editor = "Chiruzzo, Luis  and
      Ritter, Alan  and
      Wang, Lu",
    booktitle = "Proceedings of the 2025 Conference of the Nations of the Americas Chapter of the Association for Computational Linguistics: Human Language Technologies (Volume 1: Long Papers)",
    month = apr,
    year = "2025",
    address = "Albuquerque, New Mexico",
    publisher = "Association for Computational Linguistics",
    url = "https://aclanthology.org/2025.naacl-long.124/",
    doi = "10.18653/v1/2025.naacl-long.124",
    pages = "2468--2484",
    ISBN = "979-8-89176-189-6"
}

@article{adcock2026llama,
  title={The Llama 4 Herd: Architecture, Training, Evaluation, and Deployment Notes},
  author={Adcock, Aaron and Srivastava, Aayushi and Dubey, Abhimanyu and Jauhri, Abhinav and Pande, Abhinav and Pandey, Abhinav and Sharma, Abhinav and Kadian, Abhishek and Kumawat, Abhishek and Kelsey, Adam and others},
  journal={arXiv preprint arXiv:2601.11659},
  year={2026}
}

@article{singh2025openai,
  title={Openai gpt-5 system card},
  author={Singh, Aaditya and Fry, Adam and Perelman, Adam and Tart, Adam and Ganesh, Adi and El-Kishky, Ahmed and McLaughlin, Aidan and Low, Aiden and Ostrow, AJ and Ananthram, Akhila and others},
  journal={arXiv preprint arXiv:2601.03267},
  year={2025}
}

@inproceedings{ijcai2025p1155,
  title     = {Empowering LLMs with Logical Reasoning: A Comprehensive Survey},
  author    = {Cheng, Fengxiang and Li, Haoxuan and Liu, Fenrong and van Rooij, Robert and Zhang, Kun and Lin, Zhouchen},
  booktitle = {Proceedings of the Thirty-Fourth International Joint Conference on
               Artificial Intelligence, {IJCAI-25}},
  publisher = {International Joint Conferences on Artificial Intelligence Organization},
  editor    = {James Kwok},
  pages     = {10400--10408},
  year      = {2025},
  month     = {8},
  note      = {Survey Track},
  doi       = {10.24963/ijcai.2025/1155},
  url       = {https://doi.org/10.24963/ijcai.2025/1155},
}

@inproceedings{huang-chang-2023-towards,
    title = "Towards Reasoning in Large Language Models: A Survey",
    author = "Huang, Jie  and
      Chang, Kevin Chen-Chuan",
    editor = "Rogers, Anna  and
      Boyd-Graber, Jordan  and
      Okazaki, Naoaki",
    booktitle = "Findings of the Association for Computational Linguistics: ACL 2023",
    month = jul,
    year = "2023",
    address = "Toronto, Canada",
    publisher = "Association for Computational Linguistics",
    url = "https://aclanthology.org/2023.findings-acl.67/",
    doi = "10.18653/v1/2023.findings-acl.67",
    pages = "1049--1065"
}

@inproceedings{ranaldi-etal-2025-improving,
    title = "Improving Chain-of-Thought Reasoning via Quasi-Symbolic Abstractions",
    author = "Ranaldi, Leonardo  and
      Valentino, Marco  and
      Freitas, Andre",
    editor = "Che, Wanxiang  and
      Nabende, Joyce  and
      Shutova, Ekaterina  and
      Pilehvar, Mohammad Taher",
    booktitle = "Proceedings of the 63rd Annual Meeting of the Association for Computational Linguistics (Volume 1: Long Papers)",
    month = jul,
    year = "2025",
    address = "Vienna, Austria",
    publisher = "Association for Computational Linguistics",
    url = "https://aclanthology.org/2025.acl-long.843/",
    doi = "10.18653/v1/2025.acl-long.843",
    pages = "17222--17240",
    ISBN = "979-8-89176-251-0"
}

@inproceedings{
yee2024faithful,
title={Faithful and Unfaithful Error Recovery in Chain of Thought},
author={Evelyn Yee and Alice Li and Chenyu Tang and Yeon Ho Jung and Ramamohan Paturi and Leon Bergen},
booktitle={First Conference on Language Modeling},
year={2024},
url={https://openreview.net/forum?id=IPZ28ZqD4I}
}

@inproceedings{stolfo-etal-2023-mechanistic,
    title = "A Mechanistic Interpretation of Arithmetic Reasoning in Language Models using Causal Mediation Analysis",
    author = "Stolfo, Alessandro  and
      Belinkov, Yonatan  and
      Sachan, Mrinmaya",
    editor = "Bouamor, Houda  and
      Pino, Juan  and
      Bali, Kalika",
    booktitle = "Proceedings of the 2023 Conference on Empirical Methods in Natural Language Processing",
    month = dec,
    year = "2023",
    address = "Singapore",
    publisher = "Association for Computational Linguistics",
    url = "https://aclanthology.org/2023.emnlp-main.435/",
    doi = "10.18653/v1/2023.emnlp-main.435",
    pages = "7035--7052"
}

@inproceedings{
wang2023interpretability,
title={Interpretability in the Wild: a Circuit for Indirect Object Identification in {GPT}-2 Small},
author={Kevin Ro Wang and Alexandre Variengien and Arthur Conmy and Buck Shlegeris and Jacob Steinhardt},
booktitle={The Eleventh International Conference on Learning Representations },
year={2023},
url={https://openreview.net/forum?id=NpsVSN6o4ul}
}

@inproceedings{
meng2022locating,
title={Locating and Editing Factual Associations in {GPT}},
author={Kevin Meng and David Bau and Alex J Andonian and Yonatan Belinkov},
booktitle={Advances in Neural Information Processing Systems},
editor={Alice H. Oh and Alekh Agarwal and Danielle Belgrave and Kyunghyun Cho},
year={2022},
url={https://openreview.net/forum?id=-h6WAS6eE4}
}

@article{olah2020zoom,
  title={Zoom in: An introduction to circuits},
  author={Olah, Chris and Cammarata, Nick and Schubert, Ludwig and Goh, Gabriel and Petrov, Michael and Carter, Shan},
  journal={Distill},
  volume={5},
  number={3},
  pages={e00024--001},
  year={2020}
}

@inproceedings{NEURIPS2020_92650b2e,
 author = {Vig, Jesse and Gehrmann, Sebastian and Belinkov, Yonatan and Qian, Sharon and Nevo, Daniel and Singer, Yaron and Shieber, Stuart},
 booktitle = {Advances in Neural Information Processing Systems},
 editor = {H. Larochelle and M. Ranzato and R. Hadsell and M.F. Balcan and H. Lin},
 pages = {12388--12401},
 publisher = {Curran Associates, Inc.},
 title = {Investigating Gender Bias in Language Models Using Causal Mediation Analysis},
 url = {https://proceedings.neurips.cc/paper_files/paper/2020/file/92650b2e92217715fe312e6fa7b90d82-Paper.pdf},
 volume = {33},
 year = {2020}
}

@inproceedings{10.5555/2074022.2074073,
author = {Pearl, Judea},
title = {Direct and indirect effects},
year = {2001},
isbn = {1558608001},
publisher = {Morgan Kaufmann Publishers Inc.},
address = {San Francisco, CA, USA},
booktitle = {Proceedings of the Seventeenth Conference on Uncertainty in Artificial Intelligence},
pages = {411–420},
numpages = {10},
location = {Seattle, Washington},
series = {UAI'01}
}

@inproceedings{park-etal-2025-time,
    title = "Does Time Have Its Place? Temporal Heads: Where Language Models Recall Time-specific Information",
    author = "Park, Yein  and
      Yoon, Chanwoong  and
      Park, Jungwoo  and
      Jeong, Minbyul  and
      Kang, Jaewoo",
    editor = "Che, Wanxiang  and
      Nabende, Joyce  and
      Shutova, Ekaterina  and
      Pilehvar, Mohammad Taher",
    booktitle = "Proceedings of the 63rd Annual Meeting of the Association for Computational Linguistics (Volume 1: Long Papers)",
    month = jul,
    year = "2025",
    address = "Vienna, Austria",
    publisher = "Association for Computational Linguistics",
    url = "https://aclanthology.org/2025.acl-long.812/",
    doi = "10.18653/v1/2025.acl-long.812",
    pages = "16616--16643",
    ISBN = "979-8-89176-251-0"
}

@inproceedings{
hong2026a,
title={A Implies B: Circuit Analysis in {LLM}s for Propositional Logical Reasoning},
author={Guan Zhe Hong and Nishanth Dikkala and Enming Luo and Cyrus Rashtchian and Xin Wang and Rina Panigrahy},
booktitle={The Thirty-ninth Annual Conference on Neural Information Processing Systems},
year={2026},
url={https://openreview.net/forum?id=M0U8wUow8c}
}

@inproceedings{kim-etal-2025-reasoning,
    title = "Reasoning Circuits in Language Models: A Mechanistic Interpretation of Syllogistic Inference",
    author = "Kim, Geonhee  and
      Valentino, Marco  and
      Freitas, Andre",
    editor = "Che, Wanxiang  and
      Nabende, Joyce  and
      Shutova, Ekaterina  and
      Pilehvar, Mohammad Taher",
    booktitle = "Findings of the Association for Computational Linguistics: ACL 2025",
    month = jul,
    year = "2025",
    address = "Vienna, Austria",
    publisher = "Association for Computational Linguistics",
    url = "https://aclanthology.org/2025.findings-acl.525/",
    doi = "10.18653/v1/2025.findings-acl.525",
    pages = "10074--10095",
    ISBN = "979-8-89176-256-5"
}

@inproceedings{
hendrycks2021measuring,
title={Measuring Massive Multitask Language Understanding},
author={Dan Hendrycks and Collin Burns and Steven Basart and Andy Zou and Mantas Mazeika and Dawn Song and Jacob Steinhardt},
booktitle={International Conference on Learning Representations},
year={2021},
url={https://openreview.net/forum?id=d7KBjmI3GmQ}
}

@inproceedings{
todd2024function,
title={Function Vectors in Large Language Models},
author={Eric Todd and Millicent Li and Arnab Sen Sharma and Aaron Mueller and Byron C Wallace and David Bau},
booktitle={The Twelfth International Conference on Learning Representations},
year={2024},
url={https://openreview.net/forum?id=AwyxtyMwaG}
}

@article{elhage2021mathematical,
  title={A mathematical framework for transformer circuits},
  author={Elhage, Nelson and Nanda, Neel and Olsson, Catherine and Henighan, Tom and Joseph, Nicholas and Mann, Ben and Askell, Amanda and Bai, Yuntao and Chen, Anna and Conerly, Tom and others},
  journal={Transformer Circuits Thread},
  volume={1},
  number={1},
  pages={12},
  year={2021}
}

@inproceedings{
yang2025emergent,
title={Emergent Symbolic Mechanisms Support Abstract Reasoning in Large Language Models},
author={Yukang Yang and Declan Iain Campbell and Kaixuan Huang and Mengdi Wang and Jonathan D. Cohen and Taylor Whittington Webb},
booktitle={Forty-second International Conference on Machine Learning},
year={2025},
url={https://openreview.net/forum?id=y1SnRPDWx4}
}

@inproceedings{nguyen-etal-2026-improving,
    title = "Improving Chain-of-Thought for Logical Reasoning via Attention-Aware Intervention",
    author = "Nguyen, Phuong Minh  and
      Huu-Tien, Dang  and
      Inoue, Naoya",
    editor = "Demberg, Vera  and
      Inui, Kentaro  and
      Marquez, Llu{\'i}s",
    booktitle = "Findings of the {A}ssociation for {C}omputational {L}inguistics: {EACL} 2026",
    month = mar,
    year = "2026",
    address = "Rabat, Morocco",
    publisher = "Association for Computational Linguistics",
    url = "https://aclanthology.org/2026.findings-eacl.152/",
    doi = "10.18653/v1/2026.findings-eacl.152",
    pages = "2917--2941",
    ISBN = "979-8-89176-386-9"
}

\appendix


\section{AI Usage Declaration }

AI tools were used for grammar checking and formatting of tables and figures, and polish writing. All technical content
and implementations were written by the authors.
  
\section{Preliminary experiments \label{apd_prelim}}
\begin{figure}[!htbp]
    \centering  
    \includegraphics[width=.99\linewidth, keepaspectratio, 
            trim={ 0.3cm 0.3cm 0.2cm 0.2cm}, page=1, clip=true]{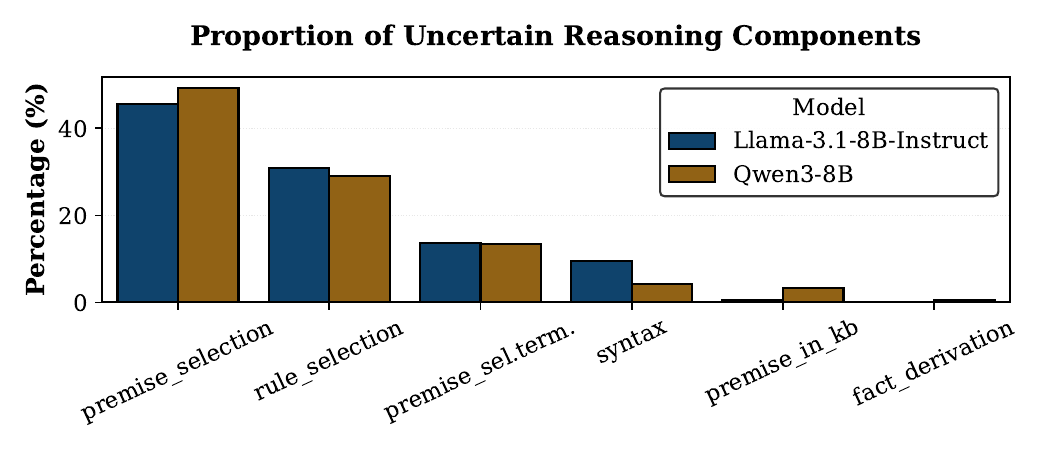} 
    \includegraphics[width=.95\linewidth, keepaspectratio, 
            trim={ 0.3cm 0.3cm 0.5cm 0.2cm}, page=1, clip=true]{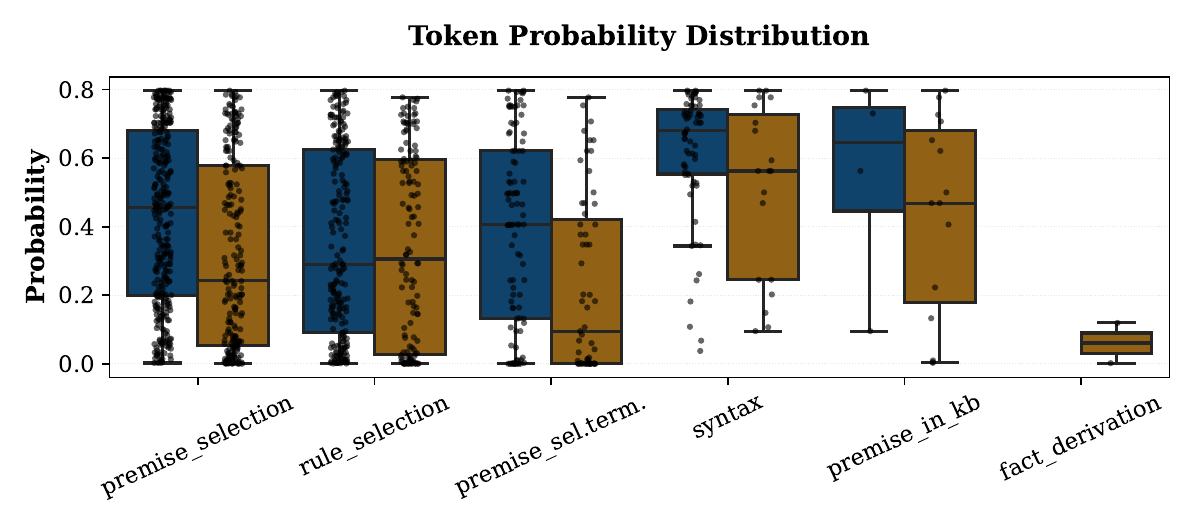} 
    \caption{Distribution of uncertain token (low token probability) grouped by reasoning component on  \texttt{Llama-3.1-8B-Instruct} and \texttt{Qwen3-8B}.}\label{fig_low_confidence_tok}
\end{figure}

We present the percentage of uncertain reasoning components and the distribution of their probabilities for the \texttt{Llama-3.1-8B-Instruct},  \texttt{Qwen3-8B}, \texttt{phi-4}, and \texttt{Qwen3-4B} models in Figure~\ref{fig_low_confidence_tok}. The experimental results show the same trend across LLMs.

\begin{figure}[!htbp]
    \centering  
    \includegraphics[width=.99\linewidth, keepaspectratio, 
            trim={ 0.3cm 0.3cm 0.2cm 0.2cm}, page=1, clip=true]{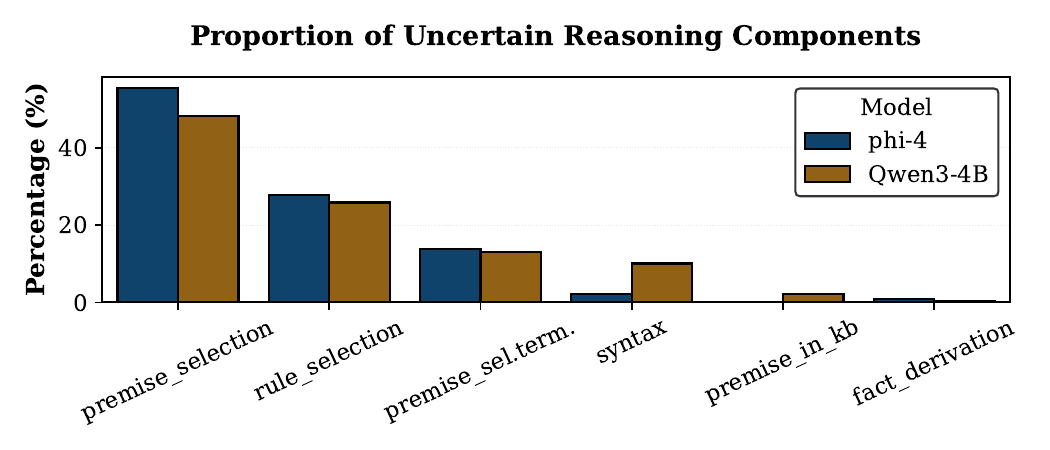} 
    \includegraphics[width=.99\linewidth, keepaspectratio, 
            trim={ 0.3cm 0.3cm 0.2cm 0.2cm}, page=1, clip=true]{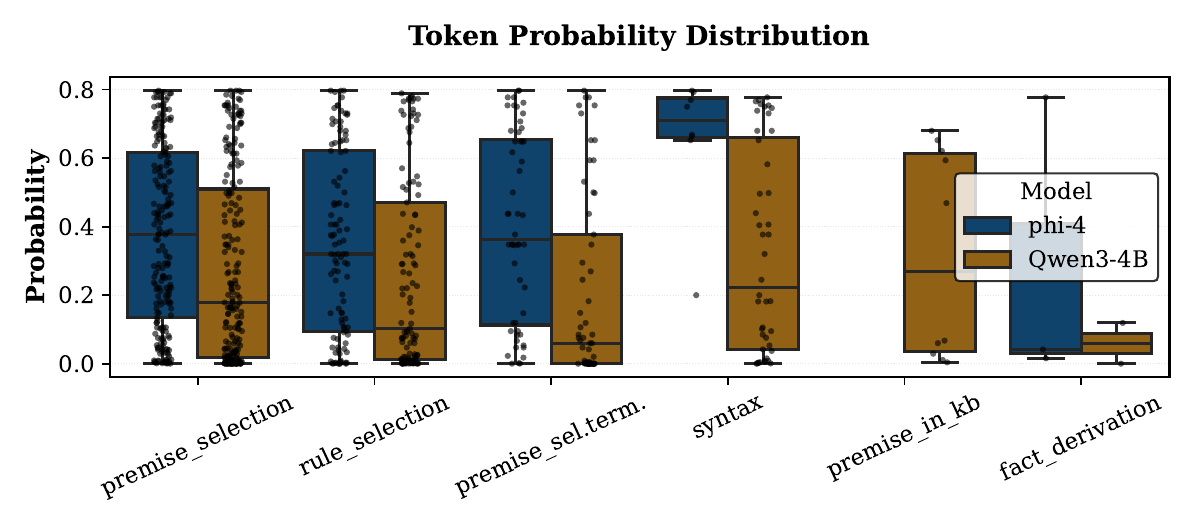} 
    \caption{Distribution of uncertain token (low token probability) grouped by reasoning component on \texttt{phi-4} and \texttt{Qwen3-4B}.}\label{fig_low_confidence_tok_apdx}
\end{figure}

\section{Data Synthesizing Algorithm \label{apd_alg_syn_data}}

    To identify reasoning circuit components, we construct a synthesized dataset $\mathcal{D}^\mathrm{pair}$ of structurally identical prompt pairs: clean prompts $p$ and corrupted prompts $\tilde{p}$ (Algorithm~\ref{alg:corrupt_generation}). Each corrupted prompt systematically modifies causal elements (e.g., fact values, rule definitions) that influence token selection at specific reasoning positions, such as premise selection or rule application. This controlled perturbation enables activation patching experiments that isolate component-level causal contributions to reasoning behavior. 
    %
    \begin{algorithm}[htbp]
        \small 
        \caption{Corrupted Prompt Generation} \label{alg:corrupt_generation}
        \begin{algorithmic}[1]
        \REQUIRE $n$ data points, $k$ demonstrations, corrupt type $c$
        \ENSURE Dataset $\mathcal{D}^{\mathrm{pair}(c)} = \{(p_i, \tilde{p}_i)\}_{i=1}^n$ \\[0.15cm]
        
        \WHILE{$i < 10 \times n$ \AND  $|\mathcal{D}| \leq n$ }
            \FOR{$z = 0$ to $k+1$}
                \STATE  \textcolor{RoyalPurple}{\textit{// Generate k-shots (problem and reasoning chain)}}
                \STATE  $(x_z, y_z) \leftarrow  \textsc{GenerateLRProblem} ()$ 
            \ENDFOR
            
            \STATE $p_i \leftarrow \textsc{Format}(\{(x_z, y_z)\}_{z=0}^{k+1})$\\[0.15cm]
            
            \STATE \textcolor{RoyalPurple}{\textit{// Corrupt clean prompt}}
            \STATE $\tilde{p}_i \leftarrow \textsc{Corrupt}(p_i, c)$\\[0.15cm]
    
            \STATE  \textcolor{RoyalPurple}{\textit{// Validate $p_i$ and $\tilde{p}_i$ share identical structure}}
            \IF{$\textsc{ValidateStructure}(p_i, \tilde{p}_i )$}
                \STATE $\mathcal{D}^{\mathrm{pair}(c)} \leftarrow \mathcal{D}^{\mathrm{pair}(c)} \cup \{(p_i, \tilde{p}_i)\}$
            \ENDIF
            \STATE $i \leftarrow  i + 1$
        \ENDWHILE
        \RETURN $\mathcal{D}^{\mathrm{pair}(c)}$
        \end{algorithmic}
    \end{algorithm}
    

\section{Circuit Discovering Results \label{apd_circuits_discovery}}
\begin{figure*}[!htbp]
    \centering  
    \includegraphics[width=1.\linewidth, keepaspectratio, 
            trim={  0.1cm 0 8.5cm 0}, page=1, clip=true]{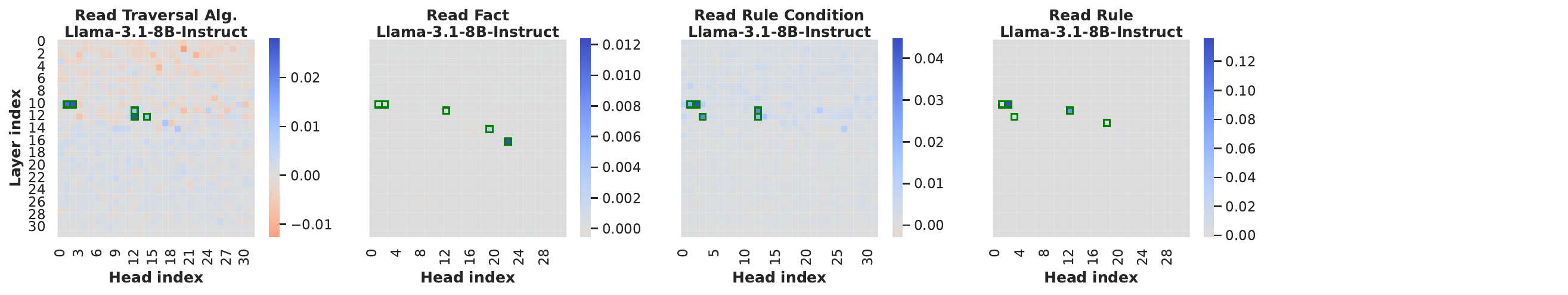}  
    \includegraphics[width=1.\linewidth, keepaspectratio, 
            trim={ 0.1cm 0 8.5cm 0}, page=1, clip=true]{images/Llama-3.1-8B-Instruct__aie_dist.pdf}  
    \includegraphics[width=1\linewidth, keepaspectratio, 
            trim={0.15cm 0.45cm 7.33cm 0.89cm}, page=1, clip=true]{images/Llama-3.1-8B-Instruct_layer_role_score.pdf}
    \caption{Distribution of AIE scores across layers and attention heads on \texttt{Llama-3.1-8B-Instruct} model. In the heatmap, the five highest-scoring heads are highlighted with green rectangles. The head roles in the line chart are aligned with the second row heat map; each layer score is computed by averaging the AIE scores of the top 15\% highest-scoring heads in that layer; the red dashed line show the layer most responsible for rule selection.}\label{fig_apd_llama3_8b__aie_distdist}
    \end{figure*}

\begin{figure*}[!htbp]
    \centering   
    \includegraphics[width=1\linewidth, keepaspectratio, 
            trim={0.1cm 0 8.5cm 0}, page=1, clip=true]{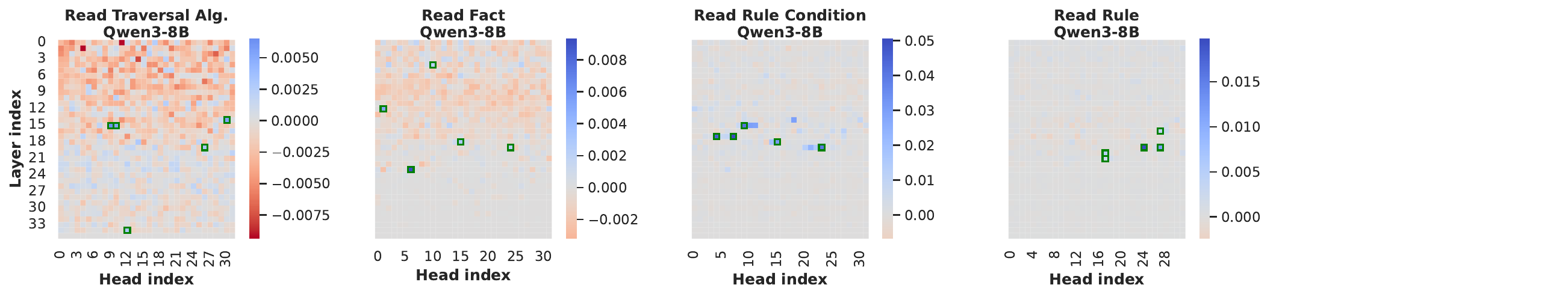}  
    \includegraphics[width=1.\linewidth, keepaspectratio, 
            trim={0.1cm 0 8.5cm 0}, page=1, clip=true]{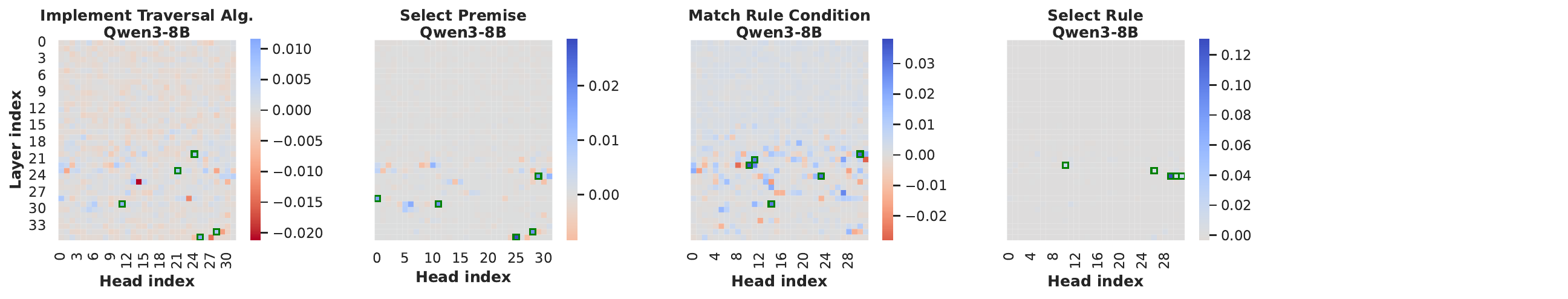}  
    \includegraphics[width=1\linewidth, keepaspectratio, 
            trim={0.15cm 0.45cm 7.33cm 0.89cm}, page=1, clip=true]{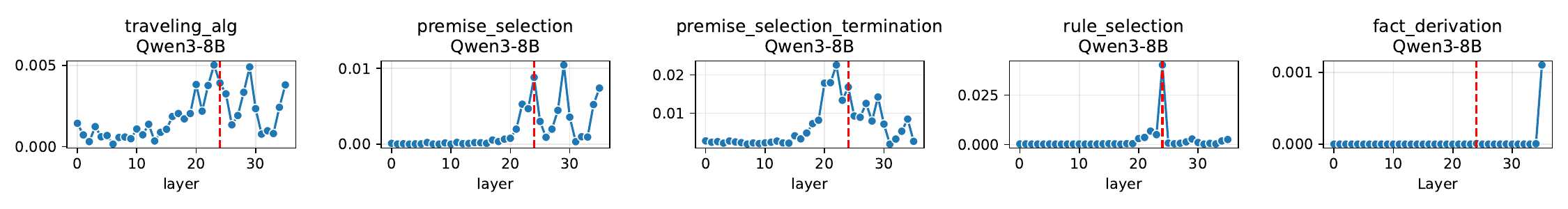}
    \caption{ Distribution of AIE scores across layers and attention heads on \texttt{Qwen3-8B} model.}\label{fig_Qwen3_8B__aie_distdist}
    \end{figure*}

\begin{figure*}[!htbp]
    \centering  
    \includegraphics[width=1\linewidth, keepaspectratio, 
            trim={0.1cm 0 8.5cm 0}, page=1, clip=true]{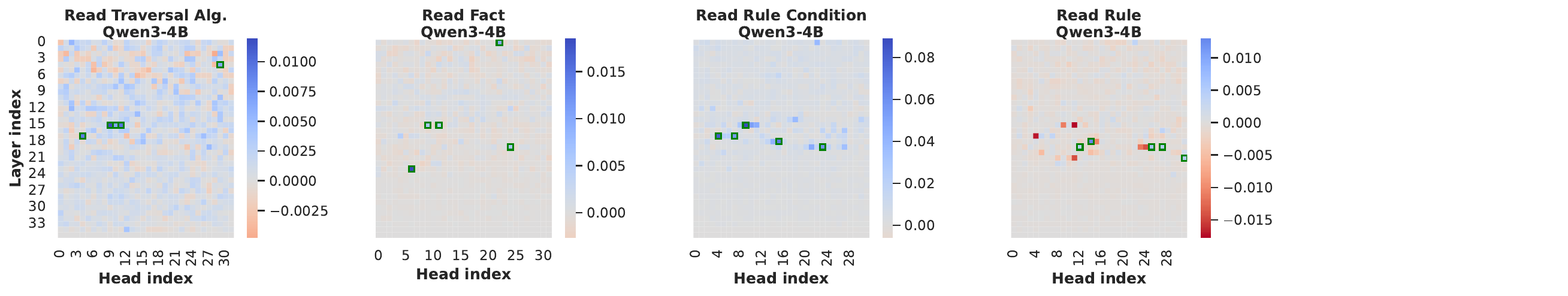}  
    \includegraphics[width=1.\linewidth, keepaspectratio, 
            trim={0.1cm 0 8.5cm 0}, page=1, clip=true]{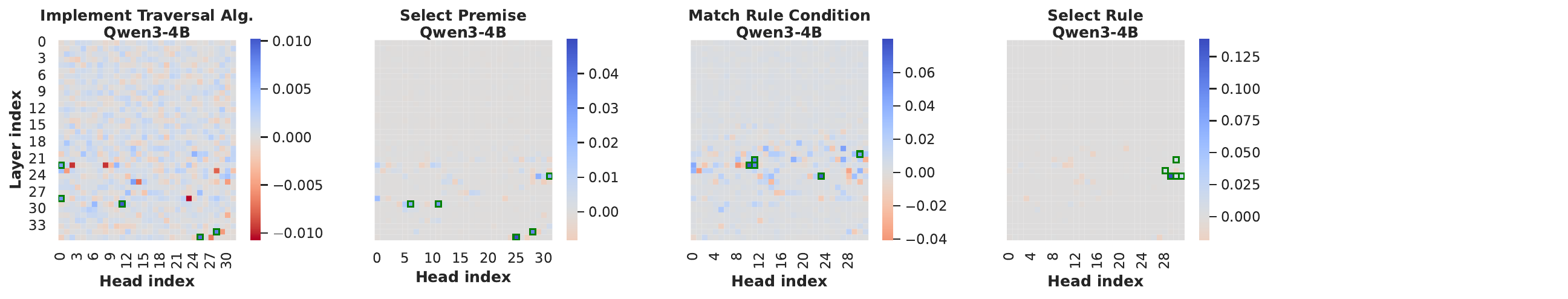}  
    \includegraphics[width=1\linewidth, keepaspectratio, 
            trim={0.15cm 0.45cm 7.33cm 0.89cm}, page=1, clip=true]{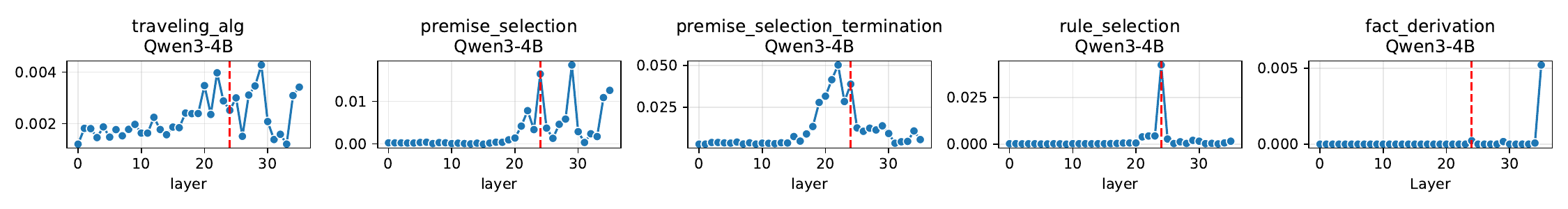}
    \caption{ Distribution of AIE scores across layers and attention heads on \texttt{Qwen3-4B} model.}\label{fig_Qwen3_4B__aie_distdist}
    \end{figure*}

\begin{figure*}[!htbp]
    \centering  
    \includegraphics[width=1\linewidth, keepaspectratio, 
            trim={ 0.cm 0.8cm 8cm 0}, page=1, clip=true]{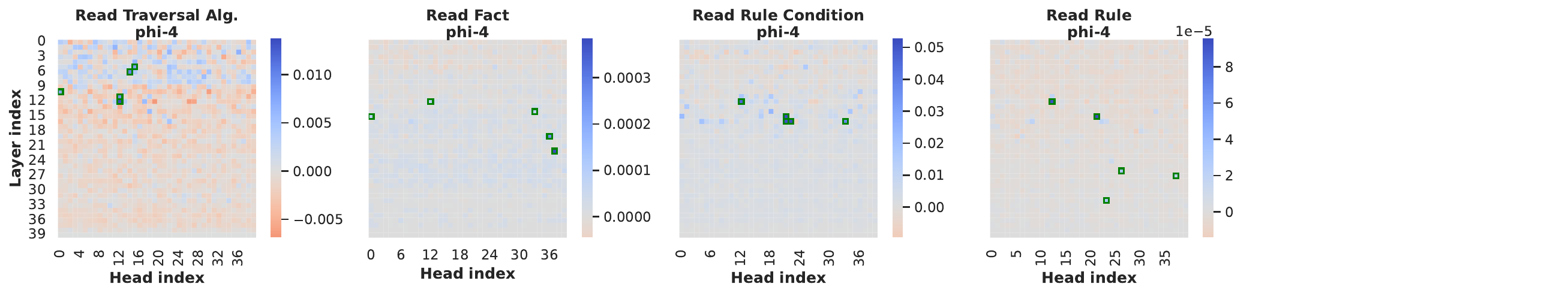} 
    \includegraphics[width=1.\linewidth, keepaspectratio, 
            trim={0.cm 0.8cm 8cm 0}, page=1, clip=true]{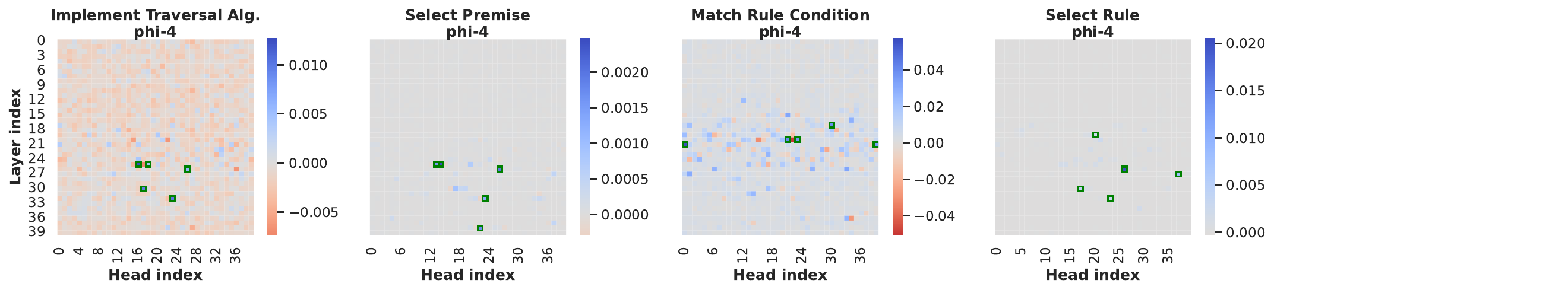}  
    \includegraphics[width=1\linewidth, keepaspectratio, 
            trim={0.15cm 0.45cm 7.33cm 0.89cm}, page=1, clip=true]{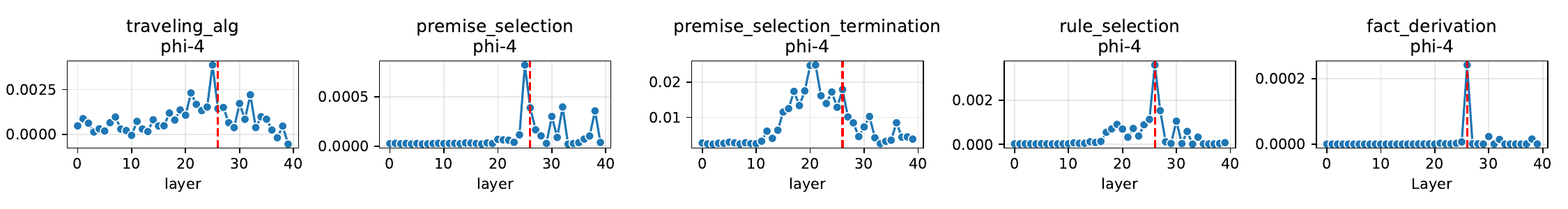}
    \caption{ Distribution of AIE scores across layers and attention heads on \texttt{phi-4} model..}\label{fig_phi_4__aie_distdist}
    \end{figure*}

\begin{figure*}[!htbp]
        \centering  
        \includegraphics[width=.88\linewidth, keepaspectratio, 
                trim={ 0 0.9cm 0 0 }, page=1, clip=true]{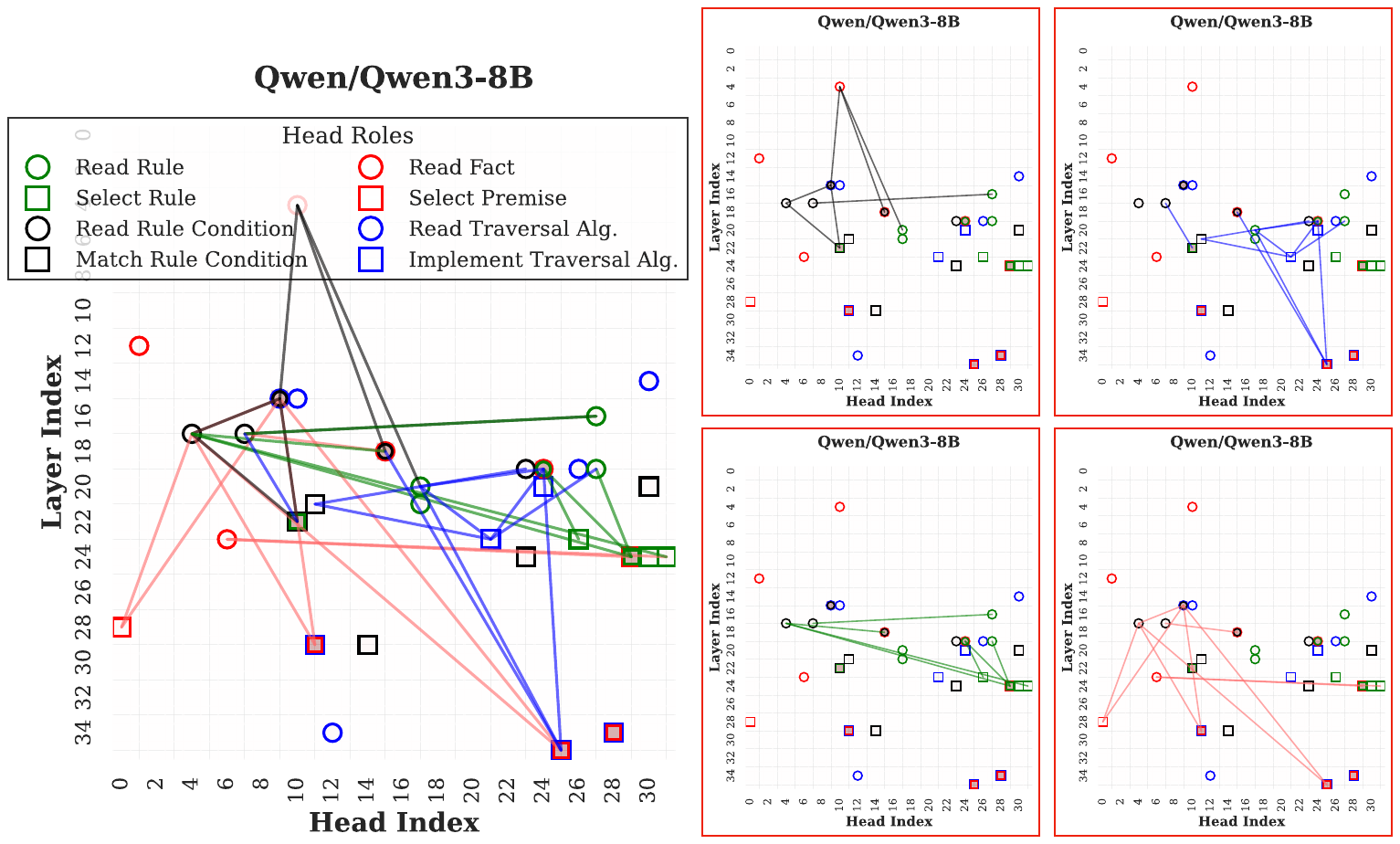}  
        \caption{\textit{Circuit network} on top-5 attention head scores associated with reasoning components in the   \texttt{Qwen3-8B} model.}\label{fig_apd_llama_qw_8b_cir}
    \end{figure*}

\begin{figure*}[!htbp]
        \centering  
        \includegraphics[width=.88\linewidth, keepaspectratio, 
                trim={ 0 0.9cm 0 0 }, page=1, clip=true]{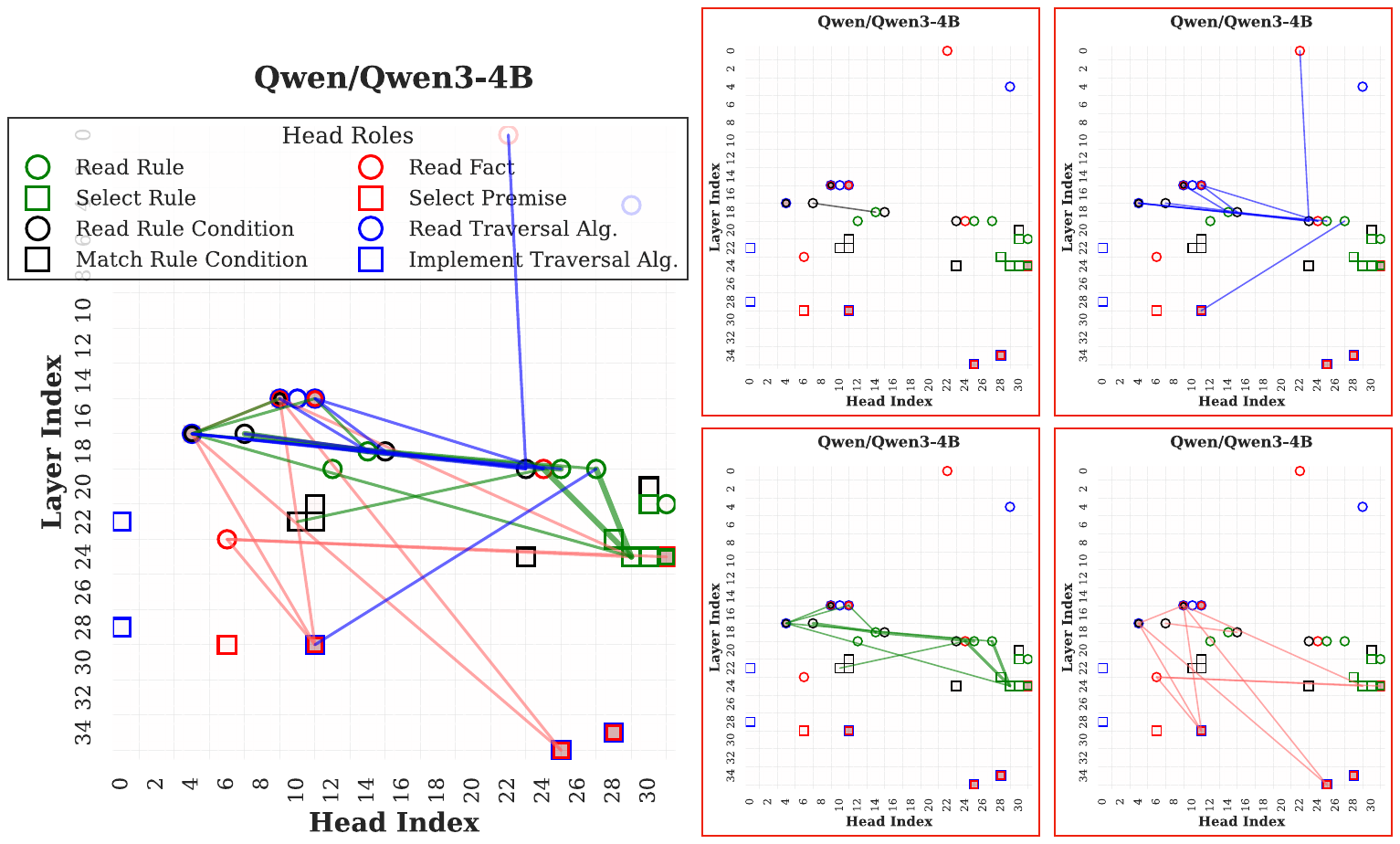}  
        \caption{\textit{Circuit network} on top-5 attention head scores associated with reasoning components in the   \texttt{Qwen3-4B} model.}\label{fig_apd_llama_qw_4b_cir}
    \end{figure*}

\begin{figure*}[!htbp]
        \centering  
        \includegraphics[width=.88\linewidth, keepaspectratio, 
                trim={ 0 0.9cm 0 0 }, page=1, clip=true]{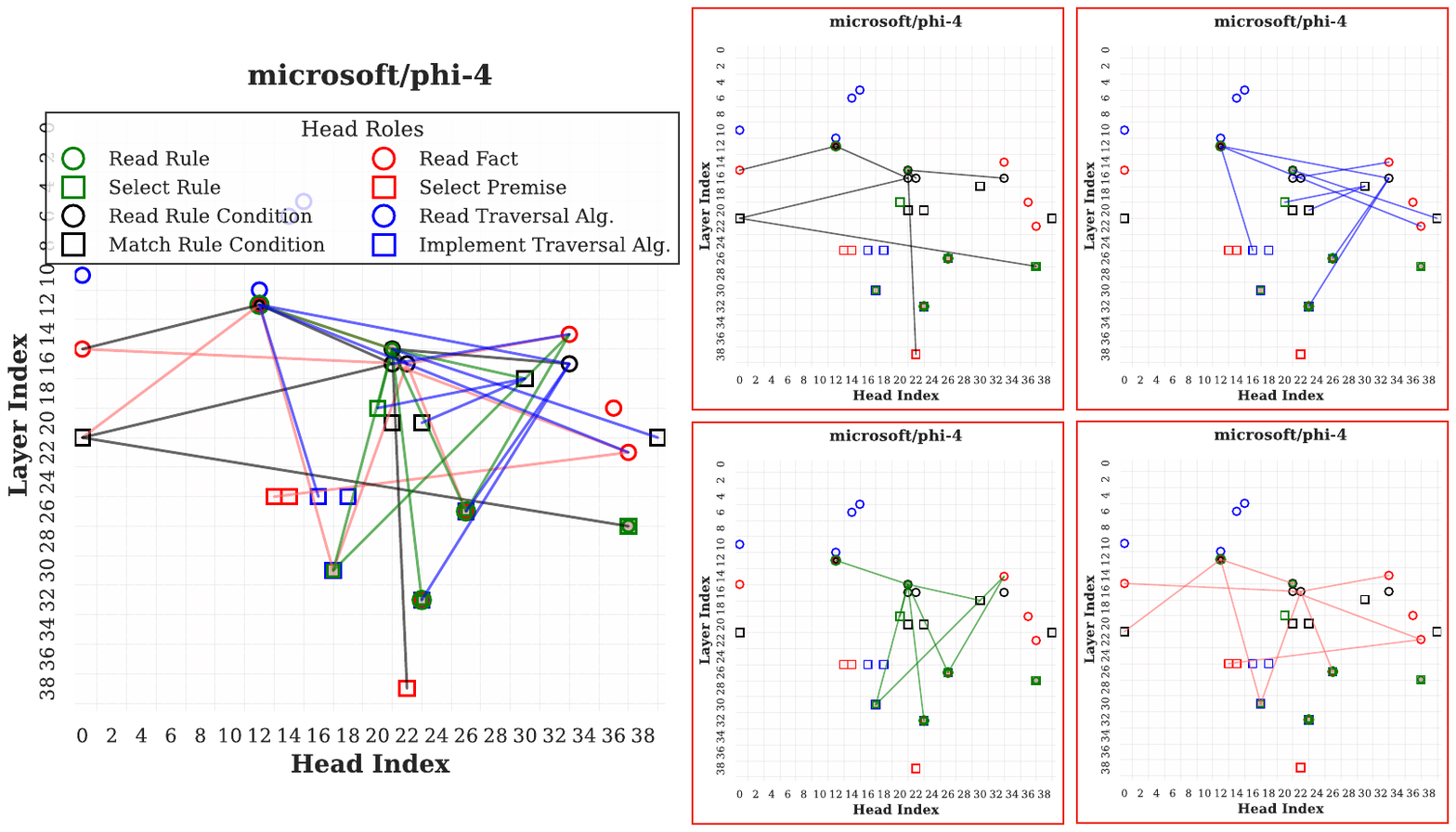}  
        \caption{\textit{Circuit network} on top-5 attention head scores associated with reasoning components in the   \texttt{phi-4} model.}\label{fig_apd_phi4_cir}
    \end{figure*}

    This section contains additional circuit discovery results across LLMs (Figures~\ref{fig_apd_llama3_8b__aie_distdist},~\ref{fig_Qwen3_8B__aie_distdist},~\ref{fig_Qwen3_4B__aie_distdist},~\ref{fig_phi_4__aie_distdist}, and \ref{fig_apd_llama_qw_8b_cir}, \ref{fig_apd_llama_qw_4b_cir},~\ref{fig_apd_phi4_cir}), which could not be fully presented in the main paper due to space limitations.

\section{Ablating LR Heads Across Datasets \label{apd_knockout_head}} 
    This section presents additional experiments on knocking out logical reasoning (LR) heads in LLMs to further validate the importance of deductive reasoning heads not only on the synthesized dataset but also on a well-known logical reasoning dataset, ProofWriter.
    \begin{figure*}[!htbp]
    \centering  
    \includegraphics[width=.99\linewidth, keepaspectratio, 
            trim={ 0.1cm 0.8cm 0.1cm 0.2cm}, page=1, clip=true]{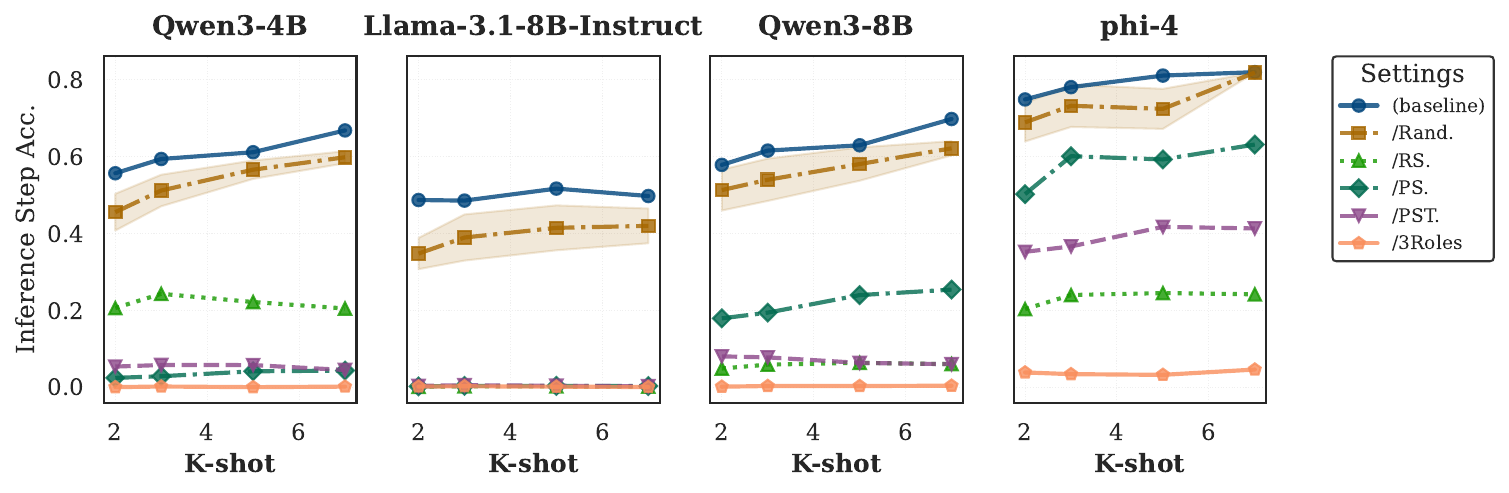} 
    \includegraphics[width=.75\linewidth, keepaspectratio, 
            trim={ 0.3cm 0.8cm 0.5cm 0.0cm}, page=1, clip=true]{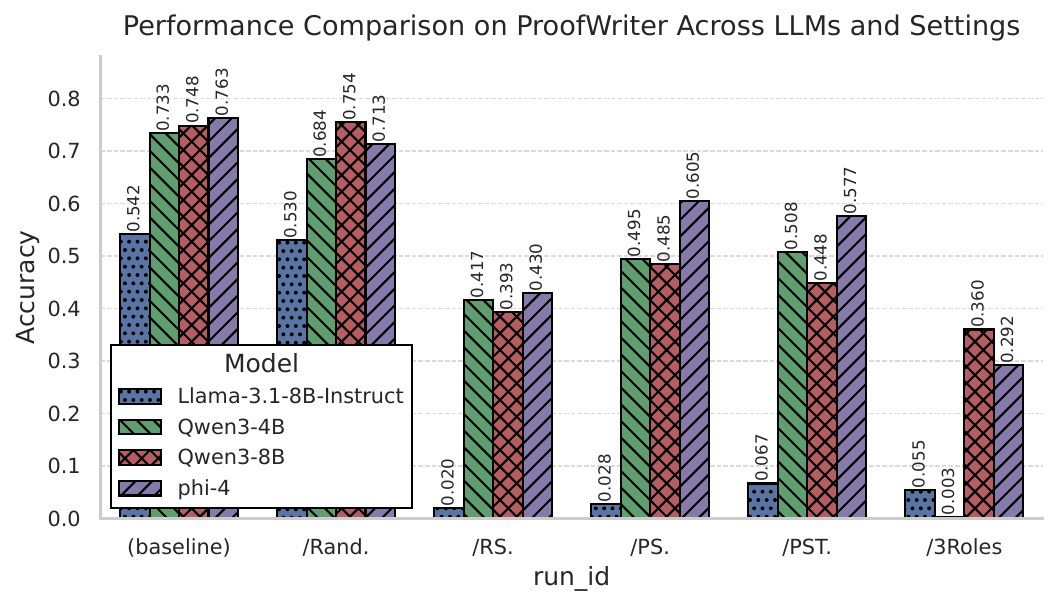} 
    \caption{Inference step accuracy when knockout top-$k$ logical reasoning heads on synthesized data ($\mathcal{D}^{\mathrm{syn}}$) and on Logical Reasoning task, ProofWriter dataset ($k=8$ with \texttt{phi-4} model and $k=5$ with other models).  }\label{fig_apd_knockout_syn_acc}
    \end{figure*}


\section{Synthesize Data Example   \label{apd_synthesized_data}}
    For a better understanding of the synthesis process, this section presents examples of clean-corrupt prompt pairs in $\mathcal{D}^{\mathrm{pair}(c)}$ across different corruption types ($c$), as shown in Tables~\ref{tab_example_corrupt_PS},\ref{tab_example_corrupt_PST},\ref{tab_example_corrupt_RS}, and~\ref{tab_example_corrupt_alg}.
    
    \begin{table}[!htbp]
    \small
    \centering
    \caption{Example of a corrupt prompt type: \textit{modifying the facts} affects \textit{premise selection}. The \textcolor{blue}{blue} text represents the content of the causal span (as well as the corrupted positions) in the clean prompt, while the \textcolor{red}{red} text represents the corrupted content, and the \textcolor{gray}{gray} text will be ignored. The last corrupted position corresponds to the reasoning component, where we inspect the logit difference under changes in the causal spans (causal positions).  
    \label{tab_example_corrupt_PS} 
    }
    \resizebox{\linewidth}{!}{%
        \begin{tabular}{|p{1.2\linewidth}|}
        \hline
        \cellcolor{gray1!1}  
        \#{}\#{}\#{} Given list of facts and rules:\newline{}\#{} (Rule1): V is true\newline{}\#{} (Rule2): S is true\newline{}\#{} (Rule3): \textcolor{blue}{Q is true}\textcolor{gray}{\textrightarrow} \textcolor{red}{K is true}\newline{}\#{} (Rule4): E is true\newline{}\#{} (Rule5): L is true\newline{}\#{} (Rule6): A is true\newline{}\#{} (Rule7): If D, J then N\newline{}\#{} (Rule8): If D then E\newline{}\#{} (Rule9): If J, K then O\newline{}\#{} (Rule10): If Q then P\newline{}\#{} (Rule11): If P, D then O\newline{}\#{} (Rule12): If K, C then J\newline{}\#{} (Rule13): If W then D\newline{}\#{} (Rule14): If K then U\newline{}\#{} (Rule15): If V then F\newline{}\#{} (Rule16): If Q then F\newline{}\#{} (Rule17): If K then R\newline{}\#{} (Rule18): If O, H then M\newline{}\#{} (Rule19): If F then U\newline{}\#{} (Rule20): If M, A then T\newline{}\#{} (Rule21): If O then S\newline{}\#{} (Question): truth value of P?\newline{}\#{} (Answer): Start from the object mentioned in the question: P\newline{}KB = \{V, S, \textcolor{blue}{Q}\textcolor{gray}{\textrightarrow} \textcolor{red}{K}, E, L, A\}\newline{}=> F(KB['\textcolor{blue}{Q}\textcolor{gray}{\textrightarrow} \textcolor{red}{K}
        \textcolor{gray}{'], Rule10) => `P`\newline{}KB = \{V, S, Q, E, L, A, P\}\newline{}=> F(KB['V'], Rule15) => `F`\newline{}KB = \{V, S, Q, E, L, A, P, F\}\newline{}=> F(KB['F'], Rule19) => `U`\newline{}KB = \{V, S, Q, E, L, A, P, F, U\}\newline{}=> Validate(KB, Question=`P`) = True.}
            \\
            \hline
        \end{tabular} 
    } 
    \end{table}

    \begin{table}[!htbp]
    \small
    \centering
    \caption{Example of a corrupt prompt type: \textit{modifying rule content} affects \textit{premise selection termination}. 
    \label{tab_example_corrupt_PST} 
    }
    \resizebox{\linewidth}{!}{%
        \begin{tabular}{|p{1.2\linewidth}|}
        \hline
        \cellcolor{gray1!1}  
        \#{}\#{}\#{} Given list of facts and rules:\newline{}\textcolor{blue}{\#{} (Rule1): If R, J then A}\textcolor{gray}{\textrightarrow} \textcolor{red}{\#{} (Rule1): If V, B then N}\newline{}\#{} (Rule2): If Q, O then P\newline{}\#{} (Rule3): If J then U\newline{}\#{} (Rule4): If K then Q\newline{}\#{} (Rule5): If D then F\newline{}\#{} (Rule6): If U then C\newline{}\#{} (Rule7): If V then U\newline{}\textcolor{blue}{\#{} (Rule8): If V then N}\textcolor{gray}{\textrightarrow} \textcolor{red}{\#{} (Rule8): If V then K}\newline{}\#{} (Rule9): If G then R\newline{}\#{} (Rule10): If U, C then V\newline{}\#{} (Rule11): If A then O\newline{}\#{} (Rule12): If F then I\newline{}\#{} (Rule13): If S then H\newline{}\#{} (Rule14): If N, M then O\newline{}\#{} (Rule15): If N, K then A\newline{}\#{} (Rule16): If V, D then R\newline{}\#{} (Rule17): L is true\newline{}\#{} (Rule18): O is true\newline{}\#{} (Rule19): H is true\newline{}\#{} (Rule20): T is true\newline{}\#{} (Rule21): U is true\newline{}\#{} (Rule22): B is true\newline{}\#{} (Question): truth value of N?\newline{}\#{} (Answer): Start from the object mentioned in the question: N\newline{}KB = \{L, O, H, T, U, B\}\newline{}=> F(KB['U'], Rule6) => `C`\newline{}KB = \{L, O, H, T, U, B, C\}\newline{}=> F(KB['U', 'C'], Rule10) => `V`\newline{}KB = \{L, O, H, T, U, B, C, V\}\newline{}=> F(KB['V\textcolor{blue}{'],}\textcolor{gray}{\textrightarrow} \textcolor{red}{',} 
        \textcolor{gray}{Rule8) => `N`\newline{}KB = \{L, O, H, T, U, B, C, V, N\}\newline{}=> Validate(KB, Question=`N`) = True.}
            \\
            \hline
        \end{tabular} 
    } 
    \end{table}

    \begin{table}[!htbp]
    \small
    \centering
    \caption{Example of a corrupt prompt type: \textit{modifying rule content} affects the \textit{rule selection} component. 
    \label{tab_example_corrupt_RS} 
    }
    \resizebox{\linewidth}{!}{%
        \begin{tabular}{|p{1.2\linewidth}|}
        \hline
        \cellcolor{gray1!1}  
        \#{}\#{}\#{} Given list of facts and rules:\newline{}\#{} (Rule1): If J, T then B\newline{}\textcolor{blue}{\#{} (Rule2): If O then U}\textcolor{gray}{\textrightarrow} \textcolor{red}{\#{} (Rule2): If P then U}\newline{}\#{} (Rule3): If C, Q then D\newline{}\#{} (Rule4): If S then W\newline{}\#{} (Rule5): If M then N\newline{}\#{} (Rule6): If O, A then F\newline{}\#{} (Rule7): If S, W then G\newline{}\#{} (Rule8): If C, I then K\newline{}\#{} (Rule9): If W then D\newline{}\#{} (Rule10): If D then W\newline{}\#{} (Rule11): If U, L then C\newline{}\#{} (Rule12): If Q, T then F\newline{}\#{} (Rule13): If R, M then B\newline{}\#{} (Rule14): O is true\newline{}\#{} (Rule15): M is true\newline{}\#{} (Rule16): J is true\newline{}\#{} (Rule17): F is true\newline{}\#{} (Question): truth value of N?\newline{}\#{} (Answer): Start from the object mentioned in the question: N\newline{}KB = \{ O, M, J, F\}\newline{}=> F(KB['\textcolor{blue}{O}\textcolor{gray}{\textrightarrow} \textcolor{red}{M}'], Rule\textcolor{blue}{2}\textcolor{gray}{\textrightarrow} \textcolor{red}{5}
        \textcolor{gray}{) => `U`\newline{}KB = \{O, M, J, F, U\}\newline{}=> F(KB['M'], Rule5) => `N`\newline{}KB = \{O, M, J, F, U, N\}\newline{}=> Validate(KB, Question=`N`) = True.}
            \\
            \hline
        \end{tabular} 
    } 
    \end{table}

    \begin{table}[!htbp]
    \small
    \centering
    \caption{Example of a corrupt prompt type: \textit{changing the traversal algorithm in demonstrations} affects the \textit{premise selection} component.
    \label{tab_example_corrupt_alg} 
    }
    \resizebox{\linewidth}{!}{%
        \begin{tabular}{|p{1.3\linewidth}|}
        \hline
        \cellcolor{gray1!1}  
        \#{}\#{}\#{} Given list of facts and rules:\newline{}\#{} (Rule1): If L then J\newline{}\#{} (Rule2): If O, S then F\newline{}\#{} (Rule3): If U then M\newline{}\#{} (Rule4): If N then L\newline{}\#{} (Rule5): If S, H then R\newline{}\#{} (Rule6): If L, I then F\newline{}\#{} (Rule7): If P then I\newline{}\#{} (Rule8): If J, A then B\newline{}\#{} (Rule9): S is true\newline{}\#{} (Rule10): N is true\newline{}\#{} (Rule11): P is true\newline{}\#{} (Question): truth value of F?\newline{}\#{} (Answer): Start from the object mentioned in the question: F\newline{}KB = \{S, N, P\}\newline{}=> F(KB['N'], Rule4) => `L`\newline{}KB = \{S, N, P, L\}\newline{}=> F(KB['\textcolor{blue}{L}\textcolor{gray}{\textrightarrow} \textcolor{red}{P}'], Rule\textcolor{blue}{1}\textcolor{gray}{\textrightarrow} \textcolor{red}{7}) => `\textcolor{blue}{J}\textcolor{gray}{\textrightarrow} \textcolor{red}{I}`\newline{}KB = \{S, N, P, L, \textcolor{blue}{J}\textcolor{gray}{\textrightarrow} \textcolor{red}{I}\}\newline{}=> F(KB['\textcolor{blue}{P}\textcolor{gray}{\textrightarrow} \textcolor{red}{L}'], Rule\textcolor{blue}{7}\textcolor{gray}{\textrightarrow} \textcolor{red}{1}) => `\textcolor{blue}{I}\textcolor{gray}{\textrightarrow} \textcolor{red}{J}`\newline{}KB = \{S, N, P, L, \textcolor{blue}{J}\textcolor{gray}{\textrightarrow} \textcolor{red}{I}, \textcolor{blue}{I}\textcolor{gray}{\textrightarrow} \textcolor{red}{J}\}\newline{}=> F(KB['L', 'I'], Rule6) => `F`\newline{}KB = \{S, N, P, L, \textcolor{blue}{J}\textcolor{gray}{\textrightarrow} \textcolor{red}{I}, \textcolor{blue}{I}\textcolor{gray}{\textrightarrow} \textcolor{red}{J}, F\}\newline{}=> Validate(KB, Question=`F`) = True.\newline{}-------\newline{}\#{}\#{}\#{} Given list of facts and rules:\newline{}\#{} (Rule1): If O then W\newline{}\#{} (Rule2): If L then B\newline{}\#{} (Rule3): If M then U\newline{}\#{} (Rule4): If B then L\newline{}\#{} (Rule5): If I then L\newline{}\#{} (Rule6): If B then V\newline{}\#{} (Rule7): If Q then V\newline{}\#{} (Rule8): If A, K then F\newline{}\#{} (Rule9): L is true\newline{}\#{} (Rule10): Q is true\newline{}\#{} (Rule11): I is true\newline{}\#{} (Question): truth value of V?\newline{}\#{} (Answer): Start from the object mentioned in the question: V\newline{}KB = \{L, Q, I\}\newline{}=> F(KB['L'], Rule2) => `B`\newline{}KB = \{L, Q, I, B\}\newline{}=> F(KB['\textcolor{blue}{B}\textcolor{gray}{\textrightarrow} \textcolor{red}{Q}'], Rule\textcolor{blue}{6}\textcolor{gray}{\textrightarrow} \textcolor{red}{7}) => `V`\newline{}KB = \{L, Q, I, B, V\}\newline{}=> Validate(KB, Question=`V`) = True.\newline{}-------\newline{}\#{}\#{}\#{} Given list of facts and rules:\newline{}\#{} (Rule1): If S then G\newline{}\#{} (Rule2): If R then O\newline{}\#{} (Rule3): If T then W\newline{}\#{} (Rule4): If A, M then I\newline{}\#{} (Rule5): If K then E\newline{}\#{} (Rule6): If E then U\newline{}\#{} (Rule7): If H then C\newline{}\#{} (Rule8): If G then V\newline{}\#{} (Rule9): If O, G then I\newline{}\#{} (Rule10): If I then C\newline{}\#{} (Rule11): If O then V\newline{}\#{} (Rule12): If W, K then N\newline{}\#{} (Rule13): If V, L then W\newline{}\#{} (Rule14): If V, F then S\newline{}\#{} (Rule15): If F, W then A\newline{}\#{} (Rule16): If Q, P then H\newline{}\#{} (Rule17): J is true\newline{}\#{} (Rule18): M is true\newline{}\#{} (Rule19): R is true\newline{}\#{} (Rule20): I is true\newline{}\#{} (Rule21): A is true\newline{}\#{} (Rule22): B is true\newline{}\#{} (Question): truth value of V?\newline{}\#{} (Answer): Start from the object mentioned in the question: V\newline{}KB = \{J, M, R, I, A, B\}\newline{}=> F(KB['R'], Rule2) => `O`\newline{}KB = \{J, M, R, I, A, B, O\}\newline{}=> F(KB['\textcolor{blue}{O}\textcolor{gray}{\textrightarrow} \textcolor{red}{I}
        \textcolor{gray}{'], Rule11) => `V`\newline{}KB = \{J, M, R, I, A, B, O, V\}\newline{}=> Validate(KB, Question=`V`) = True. }
            \\
            \hline
        \end{tabular} 
    } 
    \end{table}
\end{document}